\documentclass[11pt]{article}

\usepackage[final]{acl}

\usepackage{times} 
\usepackage{latexsym} 
\usepackage{amsmath} 
\usepackage{amssymb} 
\usepackage{booktabs} 
\usepackage{float} 
\usepackage{multirow} 
\usepackage{mdframed} 
\usepackage{soul} 
\usepackage{xcolor} 
\usepackage{fontawesome5} 
 
\definecolor{deletedred}{RGB}{255,225,225} 
\definecolor{retainedgreen}{RGB}{225,245,225} 
\definecolor{criticalblue}{RGB}{220,235,255} 
 
\sethlcolor{deletedred} 
\newcommand{\deleted}[1]{{\sethlcolor{deletedred}\hl{#1}}} 
\newcommand{\retained}[1]{{\sethlcolor{retainedgreen}\hl{#1}}} 
\newcommand{\critical}[1]{{\sethlcolor{criticalblue}\hl{#1}}} 
\sethlcolor{yellow!25} 
 
\usepackage[T1]{fontenc} 

\usepackage[utf8]{inputenc}

\usepackage{microtype}

\usepackage{inconsolata}

\usepackage{graphicx}

\title{\Large\textbf{\raisebox{-0.18\height}{\includegraphics[height=1.8em]{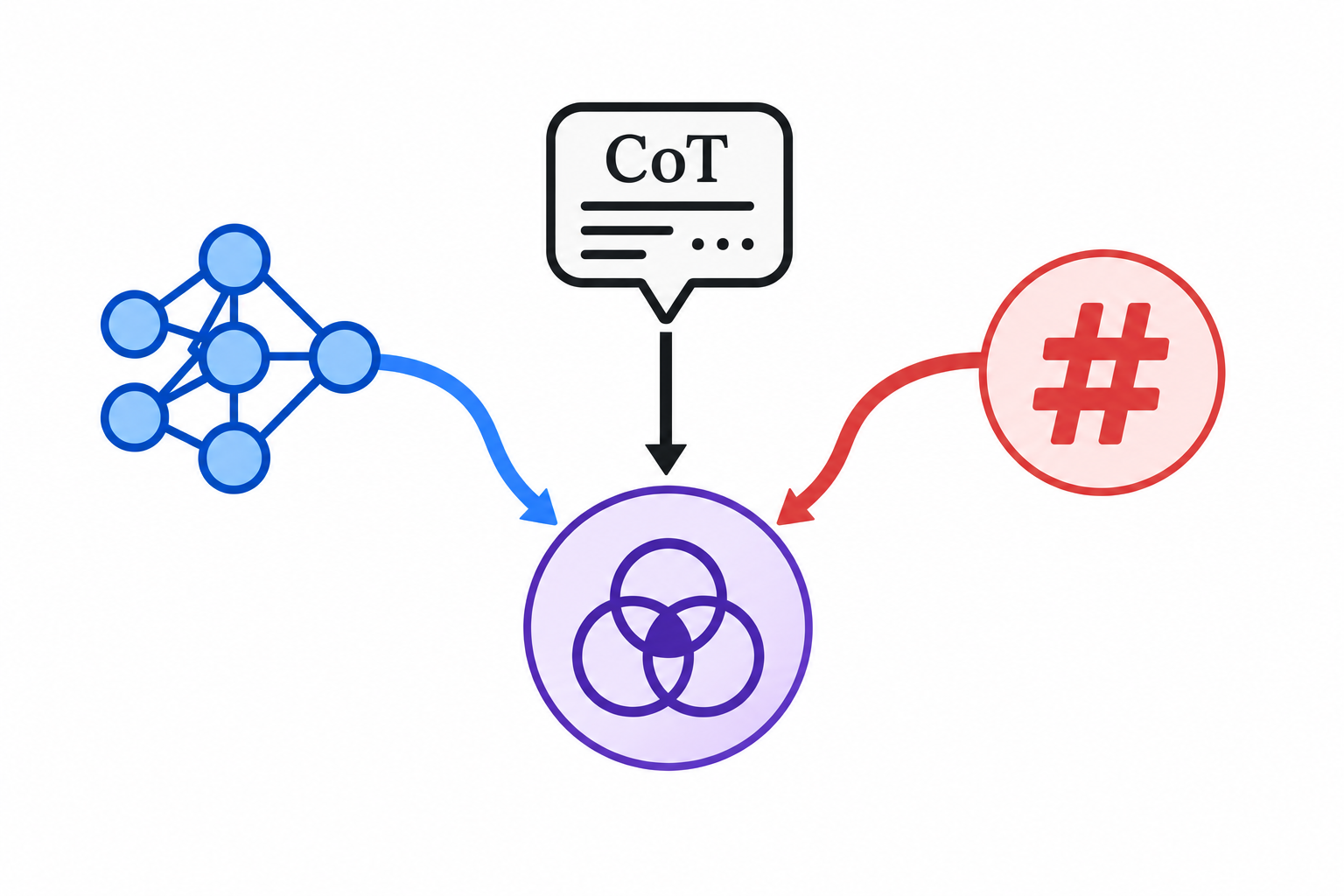}} Robust Dual-Signal Fusion: Hybrid Neuro-Symbolic Gating with
Compressed Chain-of-Thought Refinement for Irony Detection
in Social Media Texts}}

\author{
Ankit Bhattacharjee, Krityapriya Bhaumik \\
Indian Institute of Technology Kharagpur, India \\
\small \textbf{Correspondence:} 
\{\href{mailto:ankit2005@kgpian.iitkgp.ac.in}{ankit2005},
\href{mailto:ktpb124578@kgpian.iitkgp.ac.in}{ktpb124578}\}@kgpian.iitkgp.ac.in
}

\date{\today}

\begin{document}

\maketitle
\begin{center}
\small
\faGithub\ \textbf{Code:}
\url{https://github.com/SrabonGitikar/RDS-Fusion}
\end{center}

\vspace{0.2 em}

\begin{abstract}
Small-scale Large Language Models (LLMs) natively default to literal semantic interpretations, making few-shot irony detection a persistent challenge in noisy, user-generated text. In this study, we show that, despite this limitation, inference from compact LLMs can still be effectively harnessed for this task. We introduce the \textbf{Robust Dual-Signal (RDS) Fusion} framework, a hybrid neuro-symbolic architecture that utilizes a compressed Chain-of-Thought (CoT) of an LLM, alongside a static, pretrained RoBERTa and a symbolic prior module in two downstream fusions. RDS Fusion demonstrates better performance than an unrestricted reasoning ablation, while requiring substantially lower inference time. We also find that in our experiments, unrestricted reasoning does not yield better results than restricted reasoning. Evaluated on a strictly held-out TweetEval test set ($N=734$), RDS Fusion achieves $78.1\%$ accuracy and a Macro F1 of $0.777$, which is comparable to a finetuned BERTweet-base. On the heavily imbalanced iSarcasm dataset, the frozen CoT pipeline suppresses $22.5\%$ of baseline false positives, yielding a Macro F1 of $0.6726$ and Ironic F1 of $0.4821$, outperforming multiple heavily supervised SemEval transformer ensembles. Paired McNemar's tests show that, while adding the symbolic prior to the neural baseline yields an insignificant gain, and the RDS Fusion is statistically insignificant compared to the combined RoBERTa and symbolic prior ablation; the concurrent fusion achieves a statistically significant improvement over the standalone baseline ($p=0.005$).
\end{abstract}

\section{Introduction}
\label{sec:intro}

Detecting irony in unstructured social media text remains a persistent NLP challenge. Compact, open-weight LLMs natively default to literal interpretations, frequently missing the contextual inversions that define irony. In fact, standalone few-shot evaluation of \texttt{Qwen2.5-3B-Instruct} \cite{yang2024qwen25} on unstructured irony yields a poor performance (Macro F1 $<$ 0.500), suggesting that small scale local models cannot function as primary detectors in this domain without structural grounding. While Chain-of-Thought (CoT) prompting enables models to map complex pragmatics \cite{wei2022chain}, unconstrained CoT introduces severe computational latency for tasks involving chain-of-thought \cite{tokenskip2025}. Frameworks like TokenSkip \cite{tokenskip2025} reduce inference overhead via compression of the chain-of-thought, but depend on supervised finetuning (SFT) for domain adaptation and are primarily validated on structured datasets (MATH \cite{hendrycks2021math}, GSM8K \cite{cobbe2021gsm8k}). However, the application of chain-of-thought and its compression are less explored in tasks like sentiment analysis and sarcasm detection.

While fine-tuning remains the standard paradigm for adapting LLMs to narrow classification tasks, it carries practical drawbacks. The objective of our framework is not to replace supervised fine-tuning, but to bypass the necessity for it. By demonstrating that inference-time neuro-symbolic reasoning can recover fine-tuned performance levels, we establish a mechanism to achieve task-specific parity while preserving the frozen base model's capabilities.

We identify two concurrent failure modes in naive CoT compression applied to irony: (1) a \emph{literal bias} in few-shot LLMs that causes over-estimation of literal intent, and (2) a \emph{precision collapse} when explicit heuristic priors are added to maximize recall, as aggressive symbolic over-firing floods the system with false positives. Neither failure can be resolved in isolation.

We propose the \textbf{Robust Dual-Signal (RDS) Fusion} framework to resolve this tension. RDS Fusion is a hybrid neuro-symbolic architecture with three pillars: a pretrained RoBERTa \cite{liu2019roberta} component, a linguistically-motivated symbolic prior for explicit irony detection, and a prediction by an LLM. We obtain the chain-of-thought given by the LLM at a fixed budget of tokens, and further, compress the CoT post-generation, by the particular mechanism discussed in \ref{grad}. The exact role of the compression, the key difference from TokenSkip, and the motivation behind deriving a new \textit{compressed CoT based score}, when the score assigned by the LLM is not present in the compressed CoT, are discussed in Appendix \ref{app:cot}.

We empirically demonstrate that, for RDS Fusion (with post-generation pruning), the symbolic prior is load-bearing for recall while the \textbf{compressed} CoT pipeline functions primarily as a \emph{precision-recovery engine}, suppressing false positives generated by heuristic over-firing. However, an \textit{unrestricted ablation} revealed that the original LLM inference does not recover precision in the same way as the compressed RDS Fusion pipeline. Rather, the LLM frequently predicts very high scores for tweets that are literal.

\paragraph{Primary Contributions.}
(1) A few-shot neuro-symbolic fusion framework that employs a rule-based selective post-generation token pruning, without SFT. \\(2) An empirical reframing of few-shot reasoning in irony detection, showing the CoT pipeline acts primarily as a precision-restoration filter rather than a primary detector. \\(3) Validation on TweetEval ($N=734$), matching fine-tuned BERTweet-base (78.1\% accuracy, Macro F1 0.777), with statistical significance ($p=0.005$) to the baseline. \\(4) Robust few-shot cross-domain generalization on iSarcasm without parameter modification, suppressing 22.5\% of false positives and outperforming multiple supervised SemEval systems.

\section{Related Work}
\label{sec:related}

\subsection{Chain-of-Thought (CoT) Compression}
CoT prompting decomposes complex reasoning into intermediate steps \cite{wei2022chain} but introduces quadratic attention overhead \cite{vaswani2017attention} and linear KV-cache growth. Compression approaches range from perplexity-based token filtering \cite{li2023selective} to bidirectional token importance classification, as in LLMLingua-2 \cite{pan2024llmlingua2}. TokenSkip addresses this by applying quantile-thresholded compression over importance scores and utilizing Supervised Fine-Tuning (SFT) to teach LLMs to autonomously skip redundant tokens during autoregressive generation. However, the latency reduction in our framework is strictly due to the rigid threshold of \texttt{max\_new\_tokens}.

\subsection{Neuro-Symbolic and Training-Free Inference}
Inference-time compute scaling (e.g., OpenAI o1 \cite{openai2024o1}) demonstrates that dynamic reasoning budgets improve performance without parameter updates. Hybrid neuro-symbolic architectures \cite{garcez2022neurosymbolic,kautz2022third} combine neural continuous processing with symbolic discrete overrides, providing the structural grounding needed for explicit irony markers that neural pipelines smooth away \cite{riloff2013sarcasm}. RDS Fusion instantiates this paradigm for compressed CoT in social media irony.

\section{Methodology}
\label{sec:rds}

\begin{figure*}[h]
\centering

\includegraphics[width=0.9\textwidth]{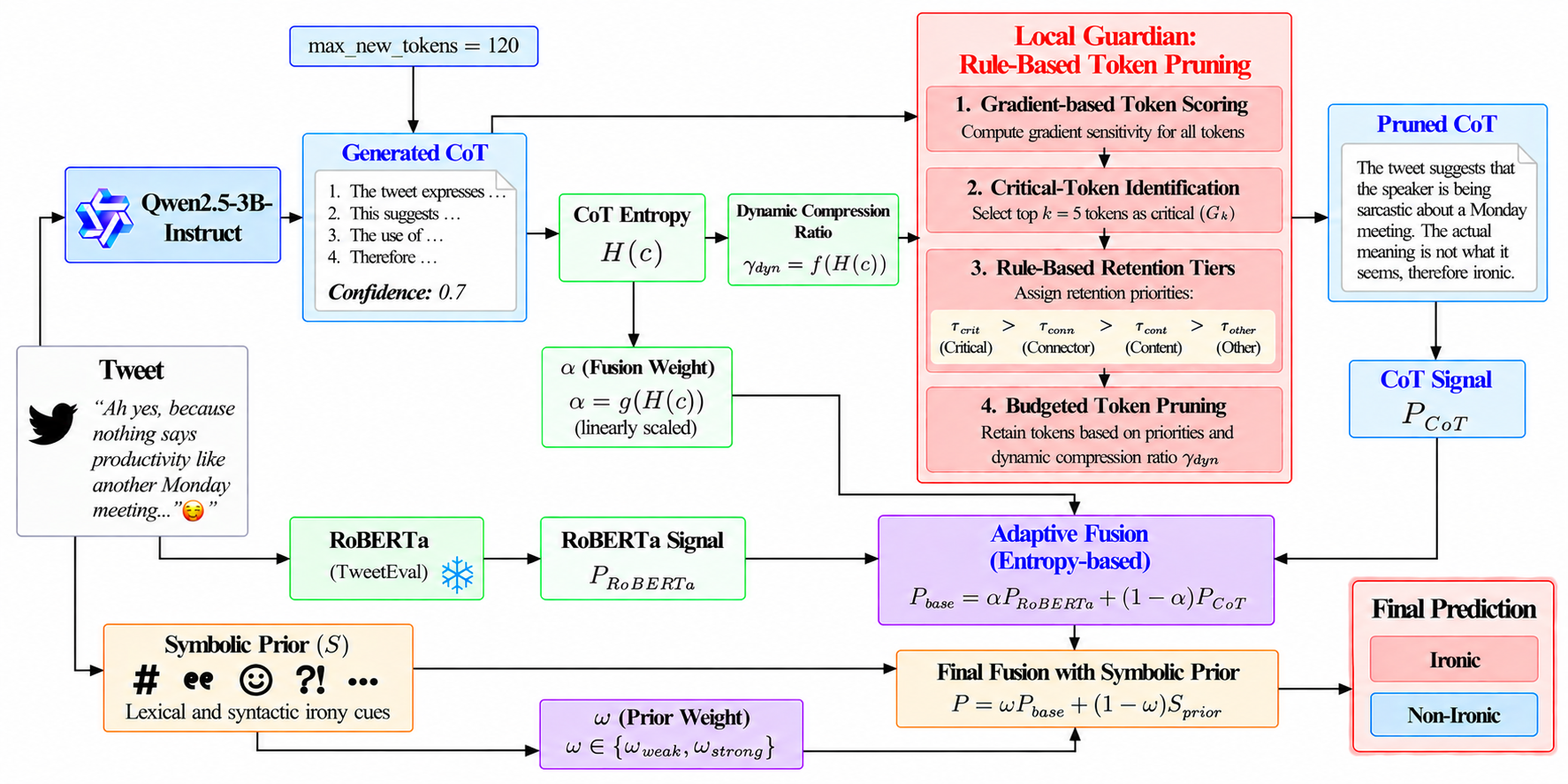}
\caption{The RDS Fusion pipeline architecture.}
\label{fig:rds_pipeline}
\end{figure*}

RDS Fusion employs a two-stage CoT compression process, consisting of a fixed generation-time token budget followed by a post-generation token pruning. Figure~\ref{fig:rds_pipeline} illustrates the full pipeline. First, the LLM is constrained to generate at most 120 new tokens. This fixed budget limits the computational cost of autoregressive reasoning.

Second, the generated trajectory is subjected to task-aware post-generation pruning. The Local Guardian assigns $k$ tokens as critical according to the gradient-based sensitivity of individual tokens to the RoBERTa classification objective. Tokens are classified into four sections, as discussed in subsection~\ref{grad}. The resulting compression therefore operates on the sequence actually generated within the fixed token budget, retaining the subset of tokens most relevant to the downstream irony decision, preserving their original order.

Together, these stages provide compression at both the generation and representation levels: the fixed generation budget bounds the length of the reasoning trajectory produced by the LLM, while post-generation pruning further reduces the information passed to the fusion layer. The resulting compressed trajectory is subsequently used to derive \(P_{\mathrm{CoT}}\), whose construction and relationship to the LLM's original confidence score are described in subsection~\ref{score}.

\subsection{Pretrained Neural Backbone}
\label{sec:neural_backbone}

The neural component of RDS Fusion employs a pretrained RoBERTa-based irony classifier as its downstream backbone. This classifier is pre-trained on the training set of TweetEval Irony \cite{barbieri2020tweeteval}. The RoBERTa parameters remain \textbf{frozen}, with the model serving as a fixed neural signal (for any possible test set) that is subsequently combined with the symbolic prior and CoT-derived signal through the RDS Fusion mechanism. Further details regarding the pretrained model, inference configuration, and implementation are provided in subsection~\ref{sec:config}.

\subsection{Few-Shot Prompt for Generation of LLM's Chain-of-Thought}
\label{subsec:prompt}
The generative pipeline relies on a manually authored 4-shot Chain-of-Thought (CoT) prompt to evaluate irony, preventing data contamination from the evaluation sets. These examples represent four canonical boundary cases: pure sincerity, explicit sarcasm, contextual ambiguity, and symbol-driven irony (e.g., usage of hashtags, emojis etc). 

The prompt structurally grounds the generator by defining irony as a pragmatic contrast and forcing the evaluation of key signals like exaggerated tone and real-world contradiction. Post-reasoning, the LLM outputs a continuous confidence score $P_{\text{CoT}}(\hat{y}=1|c) \in [0, 1]$, mapping from absolute literal ($0.0$) to ironic ($1.0$) certainty. The full prompt template is provided in appendix~\ref{app:cot_prompt}.

\subsection{Post-Generation Token Pruning Mechanism}
Given LLM $\mathcal{M}$ and CoT trajectory $c = \{c_i\}_{i=1}^m$, vanilla TokenSkip assigns importance $I(c_i)$ to each token and prunes below the $\gamma$-quantile threshold $I_\gamma = Q_\gamma(I(c_1),\dots,I(c_m))$, retaining $\tilde{c} = \{c_i \mid I(c_i) \ge I_\gamma\}$. 

\subsubsection{Prediction from Compressed CoT}
\label{score}

RDS Fusion derives its reasoning signal from the compressed trajectory rather than assuming that the LLM's original numerical confidence is preserved. Let $c$ denote the trajectory generated under the fixed
\texttt{max\_new\_tokens}=120 budget and $\tilde{c}$ its post-pruned form. We define
$P_{\mathrm{CoT}}\in[0,1]$ as the continuous irony estimate extracted from $\tilde{c}$. If an explicit score survives compression it is extracted directly; otherwise, the original score is discarded and
$P_{\mathrm{CoT}}$ is inferred from the surviving reasoning. Thus, $P_{\mathrm{CoT}}$ is a compression-derived estimator and need not equal the confidence originally expressed by the LLM. The extraction
uses explicit score/verdict matching followed by negation-aware reasoning-based heuristics, with $0.5$ denoting unresolved ambiguity and triggering a 10-token forced-verdict fallback. This process in discussed in detail in Appendix \ref{app:compression_confidence}.

\subsubsection{Predictive Entropy for Budgeting and Inference Confidence}
\label{subsec:entropy}
We compute the token-level predictive entropy, $H(c)$, to serve a dual architectural mandate: quantifying the LLM's internal confidence to regulate downstream adaptive fusion (Section \ref{fusion}), and acting as a substitute for the dynamic token allocation of Supervised Fine-Tuning (SFT).

To avoid the rigidity of compression, we introduce a dynamic ratio $\gamma_{\text{dyn}}$ calibrated via $H(c)$:
\begin{equation}
    H(c) = -\frac{1}{m} \sum_{i=1}^m \sum_{v \in \mathcal{V}} P(v|\mathbf{c}_{<i})\log P(v|\mathbf{c}_{<i}),
\end{equation}
where $m$ is the generation length, $\mathcal{V}$ is the model's vocabulary, and $\mathbf{c}_{<i}$ is the context sequence prior to step $i$. High entropy indicates structural ambiguity and therefore motivates a more aggressive compression budget. We map $H(c)$ to $\gamma_{\text{dyn}}$ via a scaled sigmoid:
\begin{equation}
    \gamma_{\text{dyn}} = \gamma_{\min} + \frac{\gamma_{\max} - \gamma_{\min}}{1 + \exp[-k(H(c) - \mu)]},
\end{equation}
where $\mu$ is the empirical mean entropy and $k$ controls sensitivity. Bounds $(\gamma_{\min}, \gamma_{\max}) = (0.1, 0.9)$ prevent degenerate compression. Unlike linear scaling, which requires hard clipping at extreme values, the sigmoid provides smooth asymptotic saturation at the tails. This structural bound ensures that extreme predictive entropy outliers—such as unconstrained hallucinations or rigidly memorized literal states—gracefully saturate at the operational limits, guaranteeing the system never enters a degenerate compression state where zero or all tokens are retained. 

Crucially, beyond budgeting, $H(c)$ explicitly measures the reliability of the few-shot inference; as detailed in Section \ref{fusion}, high predictive entropy actively regulates the fusion trust weight $\alpha$ to suppress uncertain LLM hallucinations. This $\alpha$ is heart of the neural fusion. 

\subsubsection{Selective Token Retention (Local Guardian) Module and Pruned CoT}
\label{grad}
Generation probability only measures how predictable a token is during LLM generation. It cannot determine if a token actually signals irony. Therefore, we replace probability-based scores with a task-aware piecewise-constant function to isolate the critical tokekns of the CoT. Let $\mathcal{G}_k$ denote the $k$ tokens with highest gradient norms, following gradient-based saliency approaches \cite{simonyan2013deep, li2016visualizing}, computed via backward pass through the RoBERTa classification head:
\begin{equation}
\label{grad_score}
    S(c_i) = \|\nabla_{e_{c_i}} \mathcal{L}_{\text{CE}}(\mathbf{z}, y=1)\|_2,
\end{equation}
where $\mathcal{L}_{\text{CE}}$ is the cross-entropy loss between the model logits $\mathbf{z}$ and the ironic target class ($y=1$), and $e_{c_i}$ is the $i$-th token embedding. For the entirety of this study, we have considered $k=5$. This parameter is completely adjustable.

The gradient norm in Equation \eqref{grad_score} measures the target-specific sensitivity of the RoBERTa loss to the corresponding token embedding. Mathematically it is a target-specific gradient sensitivity of the loss with respect to the token embedding. It acts as a sensitivity dial: if a specific word drastically spikes the model's irony prediction, it receives a high score, therefore enabling us to rank the tokens. This gradient-based mechanism requires access to the parameters and token embeddings of the downstream RoBERTa classifier, and therefore is not directly applicable to purely black-box classifiers or LLM APIs (e.g., GPT-4o \cite{openai2024gpt4o}, Claude 3.5 Sonnet \cite{anthropic2024claude35sonnet}) without a locally accessible differentiable model.

Let $\mathcal{C}$ denote contrastive connectors and $\mathcal{W}$ the English stopword list. The robust importance score is:
\begin{equation}
I_{\text{robust}}(c_i) =
\begin{cases}
\tau_{\text{crit}} & \text{if } c_i \in \mathcal{G}_k \\
\tau_{\text{conn}} & \text{if } c_i \in \mathcal{C} \\
\tau_{\text{cont}} & \text{if } c_i \notin \mathcal{W},\, |c_i| \ge 3 \\
\tau_{\text{low}} & \text{otherwise}.
\end{cases}
\end{equation}
In this work, we have considered $(\tau_{\text{crit}}, \tau_{\text{conn}}, \tau_{\text{cont}}, \tau_{\text{low}}) = (1000, 500, 10, 1)$. The values are chosen to draw clear boundaries between two adjacent classes of tokens. The multi-order-of-magnitude gaps ensure that gradient-critical tokens and contrastive connectors receive lexicographic retention priority whenever the compression budget admits them. The compressed trajectory is:
\begin{equation}
\label{c_RDS}
\begin{split}
    \tilde{c}_{\text{RDS}} = \text{top}_{\substack{S \subseteq c \\ |S|=\lfloor(1-\gamma_{\text{dyn}}) m\rfloor}}\, I_{\text{robust}}(c_i) \\ 
    \text{(order preserved).}
\end{split}
\end{equation}

Because $I_{\text{robust}}(c_i)$ can take only four possible values, there can be cases in which some tokens with the same importance score have to be retained while others have to be pruned. As a tie-breaking rule, the tokens are retained sequentially until the retention budget is exhausted.

\subsection{Linguistically-Motivated Prior Module}
To capture explicit authorial markers that neural pipelines smooth away, we define a symbolic prior $S_{\text{prior}}(x)$ over tiered signal sets. Let $\mathcal{R}_{\text{strong}}$ denote definitive author-labeled irony markers, $\mathcal{R}_{\text{weak}}$ soft hashtag signals, $\mathcal{R}_{\text{emoji}}$ contradictory emoji patterns, and $\mathcal{R}_{\text{elong}}$ character-elongation tokens. The full lexicon definitions appear in appendix~\ref{app:heuristics}. The prior is:
\begin{equation}
    S_{\text{prior}}(x) =
    \begin{cases}
        \rho_{\text{strong}} & \text{if } x \cap \mathcal{R}_{\text{strong}} \neq \emptyset \\
        S_{\text{comp}}(x) & \text{otherwise},
    \end{cases}
\end{equation}
where the composite weak score is capped to prevent over-firing:
\begin{equation}
\begin{aligned}
S_{\text{comp}}(x)
&= \min\!\Big(\rho_{\text{cap}},\;
\delta_w\rho_{\text{weak}} + \delta_e\rho_{\text{emoji}} \\
&\hspace{3.5em}
+ \delta_\ell\rho_{\text{elong}}\Big),
\end{aligned}
\end{equation}
where $\delta_w, \delta_e,$ and $\delta_l \in \{0, 1\}$ are binary indicator functions denoting the presence of weak irony hashtags, ironic emojis, and elongated words, respectively.
Calibrated values are in Appendix \ref{hyperparameters}.

\subsection{Entropy-Gated Adaptive Inference Fusion}
\label{fusion}
The multi-stage fusion engine balances trust among $P_{\text{CoT}}$, $P_{\text{RoBERTa}}$, and $S_{\text{prior}}$. Crucially, we utilize the predictive entropy $H(c)$ to explicitly measure the confidence of the LLM's inference, allowing the system to penalize the LLM's influence when high entropy indicates internal uncertainty.

\paragraph{Phase 1: CoT Skepticism Gate.} To prevent overconfident LLM hallucinations, extreme $P_{\text{CoT}}$ predictions are contracted toward 0.5 when $H(c) > H_{\text{sk}}$:\\
If $H(c){>}H_{\text{sk}}$ and $|P_{\text{CoT}}(\hat{y}=1|\tilde{c}_{\text{RDS}}){-}0.5| \ge 0.5{-}p_{\text{sk}}$:
\begin{equation}
\begin{split}
    &\tilde{P}_{\text{CoT}}(\hat{y}=1|\tilde{c}_{\text{RDS}}) = 0.5 \\
    &\quad + \lambda_{\text{sk}}\bigl[P_{\text{CoT}}(\hat{y}=1|\tilde{c}_{\text{RDS}}) - 0.5\bigr]; \\ 
    &\text{else} \\
    &\tilde{P}_{\text{CoT}}(\hat{y}=1|\tilde{c}_{\text{RDS}}) = P_{\text{CoT}}(\hat{y}=1|\tilde{c}_{\text{RDS}}).
\end{split}
\end{equation}

\paragraph{Phase 2: Base Fusion.} The trust weight $\alpha$ (confidence in the
RoBERTa baseline) is linearly interpolated over $[H_{\text{min}}, H_{\text{max}}]$
with bounds $[\alpha_{\text{min}}, \alpha_{\text{max}}]$. The base fusion is
then computed as:
\begin{equation}
\begin{split}
    P_{\text{base}}(\hat{y}=1|x) &= \alpha P_{\text{RoBERTa}}(\hat{y}=1|x) \\
    &\quad + (1-\alpha)\tilde{P}_{\text{CoT}}(\hat{y}=1|\tilde{c}_{\text{RDS}}).
\end{split}
\end{equation}
The calibrated hyperparameter values for the gating and fusion parameters are
provided in Appendix \ref{hyperparameters}.

\paragraph{Phase 3: Prior Injection.} The final probability interpolates the base fusion with the symbolic prior: 
\begin{equation}
    P(\hat{y}=1|x) = \omega P_{\text{base}}(\hat{y}=1|x) + (1-\omega) S_{\text{prior}}(x).
\end{equation} 
For strong signals, $\alpha$ is first capped at $\alpha_{\text{cap}}=0.30$, and a heavy blend is applied ($\omega = \omega_{\text{strong}}=0.60$). For weak signals, a moderate blend is applied ($\omega = \omega_{\text{weak}}=0.75$). If no signal is present, the system defaults to $P(\hat{y}=1|x) = P_{\text{base}}(\hat{y}=1|x)$. The final decision is $\hat{y}=1$ if $P(\hat{y}=1|x) > 0.5$.
\section{Experimental Setup}
\label{sec:setup}

\subsection{Datasets}
\paragraph{TweetEval Irony (SemEval-2018 Task 3 \cite{vanhee2018semeval})}: 784 annotated tweets serving as primary evaluation benchmark. We designate the first 50 tweets as a validation split for parameter calibration; all reported results are on the strictly held-out 734-tweet test set. For our experiments, we use the TweetEval Irony dataset exactly as distributed in its official GitHub repository, using the provided tweet texts without modification.

\paragraph{iSarcasm (SemEval-2022 Task 6, English Task A)}: 1400 self-reported sarcasm tweets ($N_{\text{ironic}}=200$, $N_{\text{literal}}=1200$). The severe class imbalance and implicit-intent distribution provide an out-of-distribution stress test. The RDS Fusion framework is deployed \emph{without any parameter modification} on this dataset.

\subsection{Models and Evaluation Pipeline}
Three components constitute the pipeline: (1) \textbf{\texttt{Qwen2.5-3B-Instruct}} generates CoT trajectories under 4-bit NF4 quantization, greedy decoding, \texttt{max\_new\_tokens}=120; (2) the \textbf{Local Guardian} classifies the tokens into four classes based on importance; (3) a post-generation compression procedure applies the resulting importance scores to construct the compressed CoT, while \textbf{cardiffnlp/twitter-roberta-base-irony} provides the discriminative baseline and gradient signals. Hardware and library details are in appendix~\ref{app:hardware}.

\subsection{Evaluated Configurations}
\label{sec:config}
Three configurations isolate each component's contribution: (i) \textbf{RoBERTa-only baseline}: task-specific fine-tuned encoder (frozen); (ii) \textbf{Ablation} (RoBERTa + Hashtag Prior): symbolic prior injected without CoT; (iii) \textbf{RDS Fusion}: complete neuro-symbolic pipeline. 

A standalone few-shot LLM configuration was evaluated but excluded from the primary tables due to a very weak performance (Macro F1 $<$ 0.500), validating its restricted role as a supplementary filter.

\section{Results and Discussion}
\label{sec:results}

\subsection{Necessity of Post-Generation Pruning}
\label{sec:pruning_necessity}

A central component of RDS Fusion is the Local Guardian, which performs post-generation pruning of the LLM reasoning trajectory before it is passed to the fusion layer. To examine whether this pruning is necessary, we compare the complete RDS Fusion framework against an otherwise identical configuration, denoted as RDS Fusion (no pruning), in which the generated CoT is passed to the fusion layer without rule-based token pruning. Both configurations use the same LLM, generation budget, symbolic prior, RoBERTa classifier, and adaptive fusion mechanism; the only difference is the application of the Local Guardian.

The results demonstrate the necessity of the pruning: RDS Fusion achieves an accuracy of \textbf{78.1\%}, compared with \textbf{75.5\%} for RDS Fusion (no pruning), yielding an absolute improvement of \textbf{2.6 percentage points}. Since the two configurations differ only in whether the Local Guardian is applied, the improvement indicates that the post-generation pruning is necessary. In this sense, it is not merely a mechanism for reducing the representation size of CoT; it acts as a task-aware filtering mechanism that improves the quality of the reasoning signal supplied to the final fusion stage.

The slightly higher ironic-class F1 of RDS Fusion (no pruning) (0.759 vs. 0.747) indicates that retaining the complete generated CoT can provide a marginal advantage for identifying ironic instances. However, this comes at the cost of lower overall performance, with RDS Fusion achieving substantially higher accuracy and Macro F1. Thus, pruning appears to trade a small amount of ironic-class F1 for a better-balanced overall classifier. All the numbers are given in table~\ref{tab:results}. It is to be noted that the run without token pruning costs marginally lower time than the one with post-generation pruning.

\subsection{System-Level Performance}
In RDS Fusion, LLM functions as a precision-restoration filter---rescuing 68 false positives (FP: $178 \to 110$) while maintaining ironic recall of 0.824. This calibration yields 78.1\% accuracy and Macro F1 of 0.777.

\begin{figure*}[h!]
\centering
\begin{minipage}{0.48\textwidth}
\centering
\includegraphics[width=\linewidth, height=6cm]{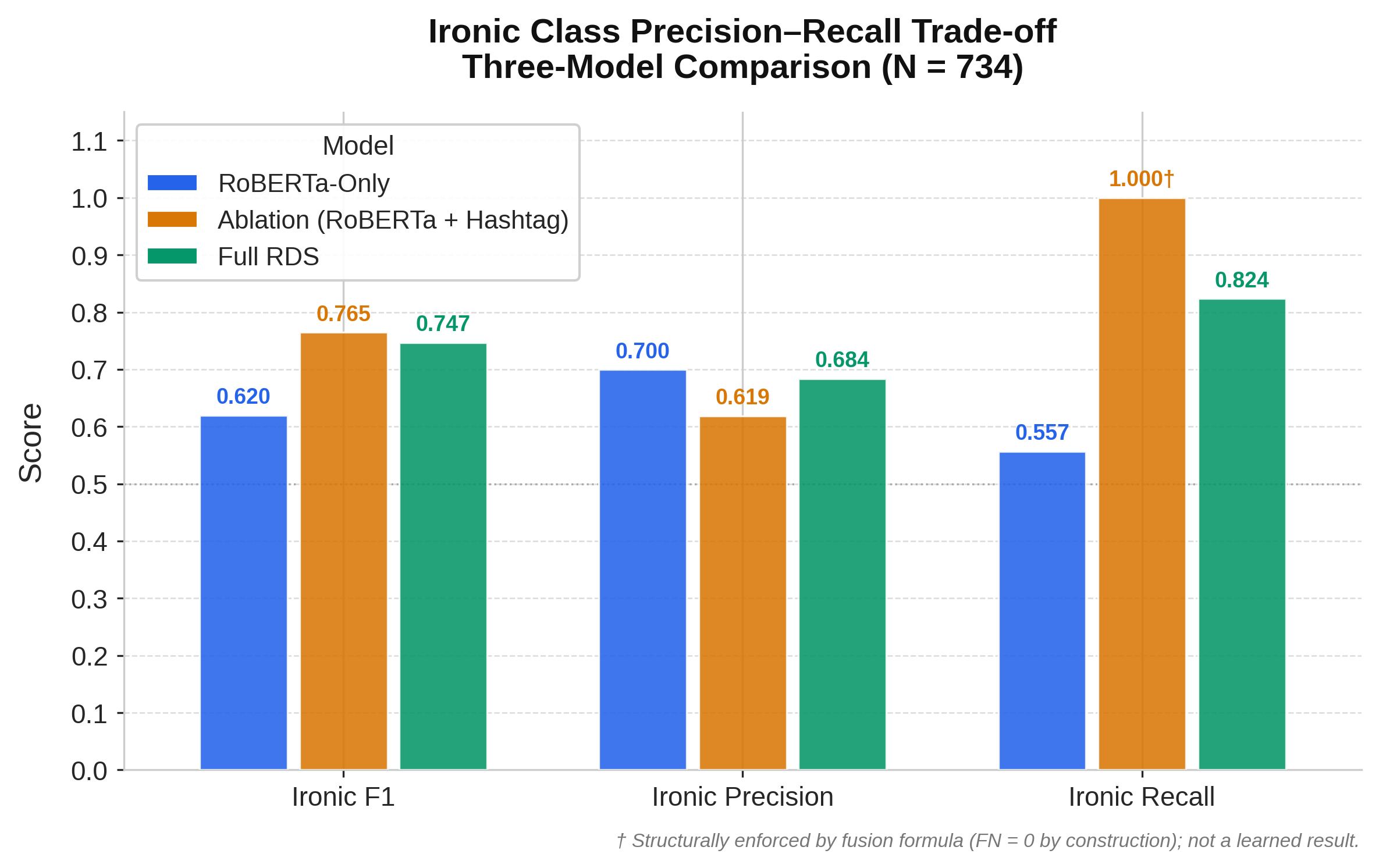}
\caption{Ironic class metrics across configurations. Ablation's perfect recall is a structural artifact forcing precision collapse; CoT restores balance.}
\label{fig:pr_tradeoff}
\end{minipage}\hfill
\begin{minipage}{0.48\textwidth}
\centering
\includegraphics[width=\linewidth, height=6cm]{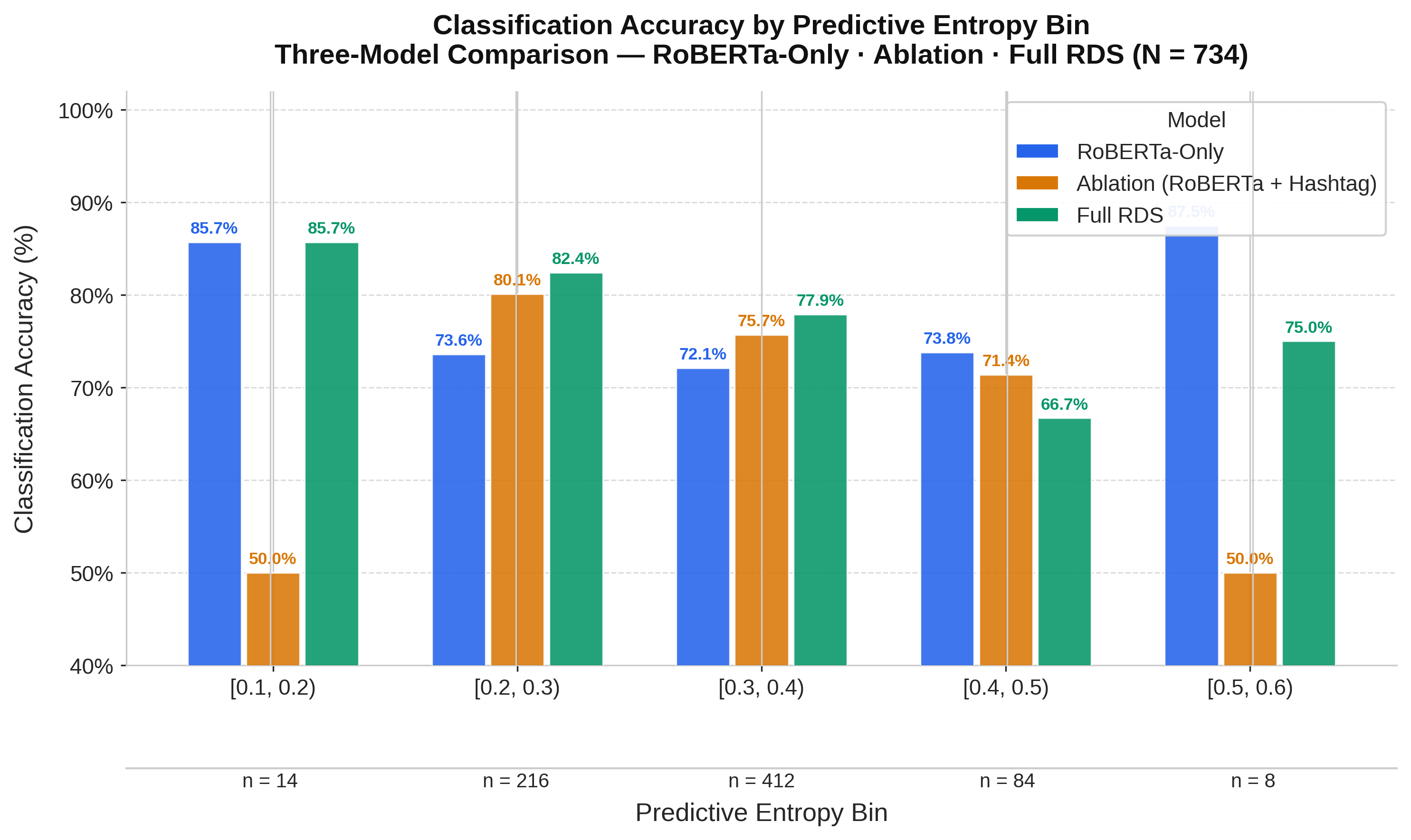}
\caption{Accuracy by entropy bin ($N=734$). Accuracy degrades monotonically with ambiguity; the ablation is uncorrelated with entropy.}
\label{fig:entropy_accuracy}
\end{minipage}

\vspace{0.5cm} 

\includegraphics[width=\textwidth, height=6cm]{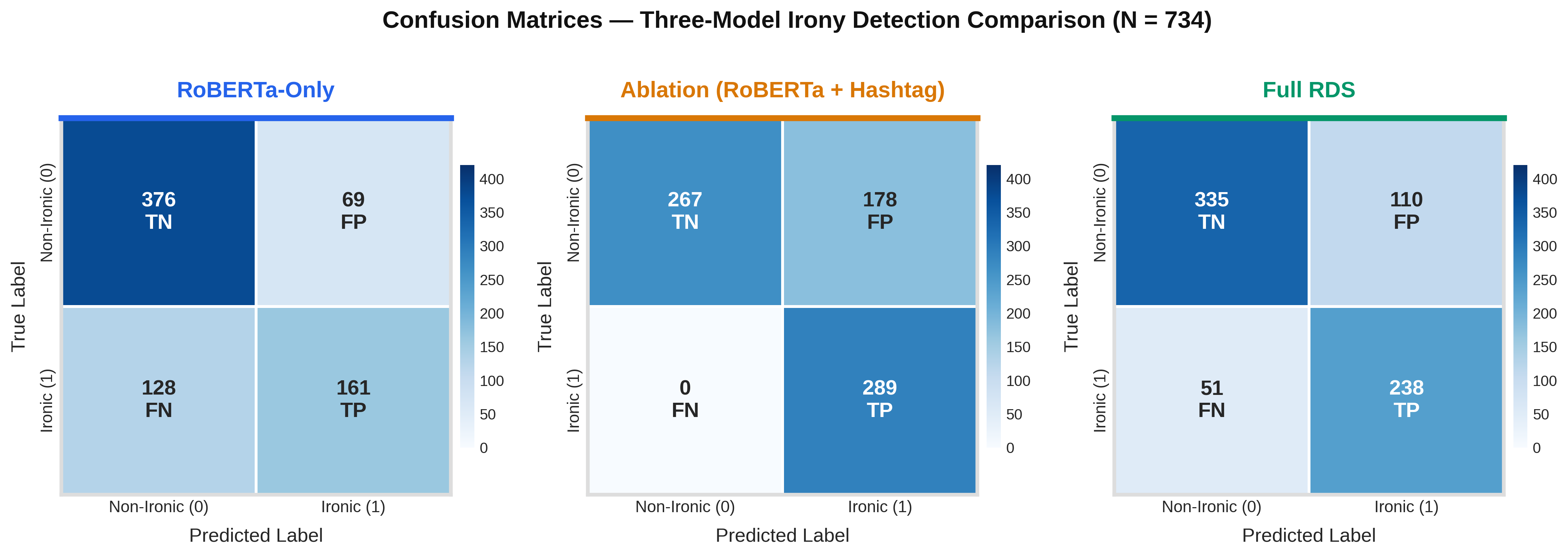}
\caption{Confusion matrices for all three TweetEval configurations. The RoBERTa baseline is conservative (FP=69, FN=127); the ablation enforces zero false negatives at the cost of FP=178; RDS Fusion fusion reduces FP to 110 while keeping FN=51.}
\label{fig:confusion_matrices}
\end{figure*}

The structural necessity of the Chain-of-Thought pipeline is visually confirmed across system metrics. As illustrated in Figure \ref{fig:pr_tradeoff} and the corresponding confusion matrices (Figure \ref{fig:confusion_matrices}), the symbolic ablation artificially maximizes ironic recall at the cost of a severe precision collapse, generating 178 false positives. RDS Fusion dynamically suppresses these heuristics to recover precision, systematically reducing false positives to 110. Furthermore, this precision recovery is tightly calibrated to reasoning complexity; Figure \ref{fig:entropy_accuracy} demonstrates that accuracy degrades logically as predictive entropy increases, whereas the ablation's error distribution remains entirely uncorrelated with predictive ambiguity.

\begin{table*}[h]
\centering
\small
\resizebox{\textwidth}{!}{
\begin{tabular}{lcccccc}
\toprule
\textbf{Method} & \textbf{Accuracy (\%)} & \textbf{Macro F1} & \textbf{Ironic F1} & \textbf{Ironic Precision} & \textbf{Ironic Recall} & \textbf{Non-Ironic F1} \\
\midrule
RoBERTa-only & 73.2 & 0.706 & 0.620 & 0.700 & 0.557 & 0.792 \\
Ablation (RoBERTa + Prior) & 75.7 & 0.757 & 0.765 & 0.619 & 1.000$^\dagger$ & 0.750 \\
Unrestricted CoT Ablation & 74.9 & 0.749 & 0.755 & 0.614 & 0.979 & 0.743 \\
RDS Fusion (no pruning) & 75.5 & 0.755 & 0.759 & 0.619 & 0.979 & 0.751 \\
\textbf{RDS Fusion} & \textbf{78.1} & \textbf{0.777} & \textbf{0.747} & \textbf{0.684} & \textbf{0.824} & \textbf{0.806} \\
\bottomrule
\end{tabular}
}
\caption{Performance on TweetEval Irony test set ($N=734$). $^\dagger$Ablation's ironic recall of 1.000 is a structural artifact forcing precision collapse.}
\label{tab:results}
\end{table*}

Table~\ref{tab:results} presents the three-way comparison on the held-out TweetEval test set. The standalone RoBERTa achieves 73.2\% accuracy with a low false positive rate but suffers from ironic recall of only 0.557. Injecting the symbolic prior (ablation) raises accuracy to 75.7\% but at a critical cost: the fusion formula forces ironic recall to 1.000 (0 false negatives) because the symbolic prior activates on all gold-ironic instances in this partition, generating 178 false positives and collapsing precision to 0.619. The Macro F1 of 0.757 therefore does not fully reflect the severe precision collapse caused by the recall-saturated operating point.

\subsection{Statistical Significance}

Paired McNemar's tests across configurations were conducted to evaluate statistical significance. It was found that, adding the symbolic prior alone over the RoBERTa baseline does not yield significance ($\chi^2=1.37$, $p=0.242$)---the recall gains are neutralized by the 178 false positives. The direct comparison between RDS Fusion and ablation also falls short of significance ($\chi^2=2.08$, $p=0.149$). Only the RDS Fusion Fusion framework achieves a statistically validated improvement over the RoBERTa baseline ($\chi^2=7.95$, $p=0.005$).

\subsection{Subpopulation Analysis}
Partitioning the test set by symbolic-prior activation provides a targeted analysis of the interaction between the heuristic module and the CoT-based fusion mechanism. The prior is activated on 417 tweets, while 317 tweets receive no symbolic-prior signal. On the prior-activated subset, the RoBERTa + Prior ablation achieves an Ironic F1 of 0.829 but a severely degraded Non-Ironic F1 of 0.131, reflecting the tendency of explicit markers to over-fire on literal tweets. RDS Fusion raises Non-Ironic F1 from 0.131 to 0.508 while maintaining comparable accuracy (71.5\% vs. 71.7\%) and a high Ironic F1 of 0.801. This provides direct evidence that the CoT signal acts as a semantic safeguard against false positives introduced by the symbolic prior. 

We found that The non-triggered subset contains \textbf{no gold-ironic example} in this evaluation partition. Consequently, its Ironic F1 is undefined. This reflects the absence of positive instances in this subset, rather than a model classification outcome. 

\begin{table*}[htbp]
\centering
\small
\resizebox{\textwidth}{!}{
\begin{tabular}{llcccc}
\hline
\textbf{Subset} & \textbf{Model} & \textbf{Accuracy (\%)} & \textbf{Macro F1} & \textbf{Ironic F1} & \textbf{Non-Ironic F1} \\
\hline
\multirow{2}{*}{Triggered ($n=417$)} & Ablation (RoBERTa + Prior) & 71.5 & 0.480 & 0.829 & 0.131 \\
& \textbf{RDS Fusion} & \textbf{71.7} & \textbf{0.655} & \textbf{0.801} & \textbf{0.508} \\
\hline
\multirow{2}{*}{Non-Triggered ($n=317$)} & Ablation (RoBERTa + Prior) & 81.4 & 0.449 & --- & 0.897 \\
& \textbf{RDS Fusion} & \textbf{86.4} & \textbf{0.464} & --- & \textbf{0.927} \\
\hline
\end{tabular}
}
\caption{Full subpopulation analysis based on symbolic prior activation. CoT precision recovery elevates Non-Ironic F1 from 0.131 to 0.508 on triggered tweets.}
\label{tab:subpopulation}
\end{table*}

\subsection{Calibration Dynamics}

We observe a statistically significant negative correlation between entropy and accuracy for RDS ($r=-0.104$, $p=0.005$), versus no correlation for the ablation ($r=-0.028$, $p=0.456$). The 10.9 percentage-point accuracy spread across entropy quartiles (vs. 3.9 for the ablation) is a direct evidence that entropy actively quantifies structural ambiguity rather than merely measuring text difficulty. The $\alpha$ distribution clusters heavily at the $[0.30, 0.40]$ floor (55.4\% of inferences), confirming the system's conservative failsafe design. However, $\alpha > 0.5$ on 41.3\% of inferences demonstrates the parameter's active contribution.

While \(\gamma_{\mathrm{dyn}}\) was introduced to provide an inference-time analogue of TokenSkip's adaptive compression behavior, we report a negative result regarding its macro-level efficacy. Unlike mathematical reasoning, where generation lengths vary drastically between complex proofs and simple calculations, our prediction task yields a comparatively constrained CoT length (mean=83.7 tokens, $\sigma=5.39$, range 60--104). Consequently, $\gamma_{\text{dyn}}$ exhibits near-zero variance (mean = 0.466, $\sigma=0.013$). However, the attribution of this negative result to the uniform nature of the task has not been validated by us. This presents a room for future work. 

\subsection{Comparison with Unrestricted CoT}
\label{sec:unrestricted_comparison}

To compare RDS Fusion with unconstrained reasoning, we conducted an unrestricted CoT ablation using the same LLM and downstream classifier, but without post-generation CoT compression and with a substantially larger generation budget of 1000 tokens. On the complete 734-example evaluation set, unrestricted CoT achieved 74.9\% accuracy and an ironic-class F1 of 0.755, compared with 78.1\% accuracy and 0.747 ironic F1 for RDS Fusion. While unrestricted reasoning therefore achieves a marginally higher ironic-class F1, it does so with a markedly different precision--recall profile: its ironic precision is only 0.614, compared with 0.684 for RDS Fusion, while its recall reaches 0.979 compared with 0.824 for RDS Fusion. Thus, increasing the reasoning budget substantially improves recall but does not recover the precision lost relative to the RoBERTa baseline. In contrast, RDS Fusion provides a more balanced precision--recall trade-off while achieving higher overall accuracy and Macro F1 (0.777 vs.\ 0.749). Furthermore, RDS Fusion was observed to be more than \textbf{2.5 times faster} than the unrestricted configuration. These results indicate that substantially longer reasoning is neither necessary nor sufficient for improved overall performance; the results of the unrestricted CoT ablation are discussed in detail in appendix~\ref{app:unrestricted_reasoning}.

\subsection{Comparison with Existing Supervised Frameworks}
Table~\ref{tab:sota_comparison} shows RDS Fusion achieving comparability with BERTweet-base \cite{nguyen2020bertweet} in overall accuracy (78.1\% vs. 78.2\%) while slightly outperforming it on Ironic F1 (0.747 vs. 0.746). Furthermore, RDS Fusion achieves a dominant Macro F1 score of 0.777, substantially surpassing CCR-Net's \cite{rahman2026ccrnet} 0.708 and demonstrating balanced predictive performance across both the ironic and non-ironic classes---entirely without task-specific fine-tuning. However, the 50-tweet calibration holdout prevents a strictly identical numerical comparison, but this performance strongly suggests that well-calibrated neuro-symbolic fusion can match or exceed the output of domain-trained encoders. 

\begin{table}[h]
\centering
\small
\begin{tabular}{p{0.3\columnwidth}lccc}
\toprule
\textbf{Model} & \textbf{Paradigm} & \textbf{Acc.} & \textbf{MF1} & \textbf{Ir.F1} \\
\midrule
CCR-Net & Supervised & 70.9 & 0.708 & 0.685 \\
BERTweet-base & Supervised & \textbf{78.2} & --- & 0.746 \\
\textbf{RDS Fusion} & Hybrid & 78.1 & \textbf{0.777} & \textbf{0.747} \\
\bottomrule
\end{tabular}
\caption{Comparison with published supervised systems on the TweetEval Irony task. BERTweet-base and CCR-Net results are reported on the full 784-tweet test split; RDS Fusion is evaluated on a held-out subset of $N=734$.}
\label{tab:sota_comparison}
\end{table}

\subsection{Few-Shot Cross-Domain Robustness on iSarcasm}

Deployed without modification on iSarcasm, the RoBERTa baseline generates 315 false positives due to distribution shift. The RDS Fusion framework suppresses 22.5\% of these false positives ($315 \rightarrow 244$; McNemar's $\chi^2 = 21.44$, $p < 0.0001$), elevating Macro F1 from 0.6562 to 0.6726. Crucially, this precision recovery does not degrade recall; RDS Fusion marginally elevates the baseline's Ironic F1 ($0.4786 \rightarrow 0.4821$).

\begin{table}[h]
\centering
\small
\begin{tabular}{p{0.43\columnwidth}cc}
\toprule
\textbf{Model} & \textbf{Eval.} & \textbf{Ir.F1} \\
\midrule
stce \cite{yuan2022stce} & Supervised SOTA & 0.6052 \\
\textbf{RDS Fusion} & \textbf{Hybrid} & \textbf{0.4821} \\
Ablation (RoBERTa + Symbolic Prior) & Hybrid & 0.4793 \\
RoBERTa-only Baseline & Supervised$^\dagger$ & 0.4786 \\
SarcasmDet \cite{abdullah2022sarcasmdet} & Supervised & 0.4300 \\
UTNLP \cite{abaskohi2022utnlp} & Supervised & 0.3800 \\
FII UAIC \cite{manoleasa2022fii} & Supervised & 0.3700 \\
\bottomrule
\end{tabular}
\caption{SemEval-2022 Task 6 leaderboard (selected subset) vs. RoBERTa, the Ablation Model \& RDS Fusion. $^\dagger$RoBERTa is trained on the train set of TweetEval training set.}
\label{tab:semeval_leaderboard}
\end{table}

As shown in table~\ref{tab:semeval_leaderboard}, few-shot RDS Fusion outperforms three out of the four selected, supervised SemEval submissions. However, iSarcasm also exposes architectural limits: the symbolic prior fires on only 1.5\% of samples ($n=21$), confirming its inertia on implicit intent. On the 21 triggered instances, frozen TweetEval weights allow the LLM's literal confidence to overpower valid priors (Ironic F1: $0.889 \to 0.571$), defining the cross-domain calibration boundary. 

Despite this localized symbolic collapse, the framework's overall outperformance of in-domain SemEval models isolates the value of the generative CoT filter. By successfully mitigating the baseline's out-of-distribution hallucinations without requiring any iSarcasm-specific fine-tuning, RDS Fusion shows that post-hoc generative reasoning can effectively substitute for domain-specific retraining when tackling novel datasets.

\subsection{A Case Study: Precision Recovery on a Literal Legal Headline (Tweet 86, TweetEval Dataset) } 

\textbf{Tweet:} \textit{\#Myanmar \#men \#plead \#not \#guilty to \#murder of \#British \#tourists...}

This is a literal news headline. The author has tokenized the text using individual hashtags, including the isolated substring \texttt{\#not}. The explicit symbolic prior immediately identifies this as a strong irony signal ($S_{\text{HTag}} = 0.88$).

\paragraph{Ablation failure:} Despite RoBERTa's low baseline probability ($P_{\text{RoBERTa}}(\hat{y}=1|x) = 0.209$), the strong symbolic trigger pushes the fused probability above 0.500, yielding a false positive. This is the canonical structural artifact of the over-aggressive prior.

\paragraph{RDS Fusion correction:} The CoT pipeline parses the semantic continuity of the legal phrase ``plead not guilty'' rather than treating ``\#not'' as an isolated pragmatic inversion. During compression, the Local Guardian explicitly whitelists literal anchoring tokens (\textit{British}, \textit{indicates}) against the budgeter. The compressed CoT gives $P_{\text{CoT}}(\hat{y}=1|\tilde{c}_{\text{RDS}}) = 0.15$. Routed at $\alpha = 0.300$, this contradicts the prior and pulls the fused probability to $P(\hat{y}=1|x) = 0.453$, yielding a correct Non-Ironic prediction. This interaction demonstrates exactly how the CoT engine functions as a semantic safeguard against symbolic over-firing.

Two more case studies are provided in appendix~\ref{app:case_study}.


\section{Conclusion}
We presented Robust Dual-Signal (RDS) Fusion, a training-free-at-inference hybrid neuro-symbolic inference pipeline achieving few-shot irony detection at a performance level comparable to supervised systems. On the held-out TweetEval test ($N=734$), RDS Fusion achieves numerical comparability with BERTweet-base and significantly improves upon the frozen, fine-tuned RoBERTa baseline ($p=0.005$). Without modification, it also surpasses multiple supervised SemEval systems on the cross-domain iSarcasm benchmark. Our findings suggest that unrestricted reasoning is not always better for a prediction task, specifically when a compact LLM is used. We also found that, the compressed (pruned) CoT in RDS Fusion functions as a \textit{precision-recovery} engine. However, neither the unrestricted CoT ablation and nor the ablation with no post-generation pruning showed this behaviour---this was solely observed for the constrained and post-pruned CoT component. These results show that calibrated frameworks requiring no additional task-specific fine-tuning of the LLM or fusion components can match or even exceed the performance of supervised frameworks.

\section{Limitations}
\label{sec:limitations}

\paragraph{Inference Latency.} The framework incurs a multi-order-of-magnitude latency penalty compared to the baseline: the RoBERTa baseline processes 1400 samples in $<$1 minute; RDS Fusion requires $\approx$5 hours ($\approx$12.86 sec/sample) due to sequential autoregressive generation by the LLM and per-sample backward passes. RDS Fusion is appropriate only in cases where supervised finetuning is not feasible due to time or computational costs. However, the framework is still a number of times faster than one that utilizes the whole, unrestricted and uncompressed CoT.

\paragraph{LLM Inference Variability.}
The LLM inference process is not guaranteed to produce an identical CoT signal for the same tweet across separate runs, making the resulting reasoning signal potentially run-dependent. Consequently, the CoT-derived component may vary across repeated evaluations even when the input tweet remains unchanged. Therefore, the overall result also may demonstrate small variations across different runs. 

\paragraph{Nearly Static Dynamic Compression Ratio.} We introduced the dynamic compression ratio so that tokens could be pruned depending on the particular CoT. However, it was found that $\gamma_{\text{dyn}}$ was almost always in the range of $0.45 - 0.51$. The near-zero variance of the dynamic retention budget ($\gamma_{\text{dyn}}$) is a negative result driven likely by the uniform nature of the task of prediction. The semantic limitations of the compact 3B-parameter model may also have compounded this inertia, however, it remains an area of future validation. 

\paragraph{Limitation of the Symbolic Prior.} On datasets where irony is primarily psychological or semantic rather than syntactic (e.g., iSarcasm), the symbolic prior remains largely dormant, activating on only very few of the samples (1.5\% in case of iSarcasm). In cases where there are literally no explicit symbolic signals, RDS Fusion becomes a purely neural cofiguration. 

\paragraph{Static Fusion Weight Fragility.} Calibration thresholds optimized on a 50-tweet TweetEval split cannot universally transfer across shifting linguistic distributions. On iSarcasm's triggered subset, frozen blending weights allowed the LLM's literal bias to overpower the valid symbolic prior, degrading Ironic F1 from 0.889 to 0.571.


\vspace{0.5em}

\section*{Acknowledgements}
We sincerely thank Professor Sourangshu Bhattacharya for his guidance as the instructor of CS60050, where the idea of RDS Fusion originated as a course project. We also thank Vaishnovi Arun for mentoring our project group, and Yuvraj Veer and Maitreyee Chakraborty for their contributions during the project. We are grateful to Professor Rajaram Lakkaraju and Professor Niloy Ganguly for their support following the acceptance of this work.

\section*{Ethics Statement}
This work uses publicly available social media datasets, \href{https://github.com/cardiffnlp/tweeteval}{\texttt{TweetEval}} and \href{https://github.com/dmbavkar/iSarcasm}{\texttt{iSarcasm}}, for research on automatic irony detection. We do not collect new user data or conduct experiments involving human participants. The datasets may contain offensive or sensitive language inherent to social media. The complete RDS Fusion implementation and generation logs are available in its \href{https://github.com/SrabonGitikar/RDS-Fusion}{\texttt{GitHub}} repository, and the \href{https://huggingface.co/cardiffnlp/twitter-roberta-base-irony}{\texttt{twitter-roberta-base-irony}} model is publicly available via Hugging Face.

\vspace{0.5em}

\bibliography{references}


\appendix

\section{Extended Neuro-Symbolic Heuristics \& Lexicons}
\label{app:heuristics}

\subsection{Contrastive Connector Vocabulary ($\mathcal{C}$)}

The contrastive connector set used in the Local Guardian is: \\
\begin{equation*}
\begin{split}
\mathcal{C} = \{&\textit{but, yet, though, although, however, despite, } \\
&\textit{whereas, still, even, never, not, no, without,} \\
&\textit{barely}\}.
\end{split}
\end{equation*}
These tokens are disproportionately responsible for carrying ironic inversion in natural language \cite{joshi2015harnessing,reyes2013multidimensional} and receive lexicographic retention priority for any compression budget, by virtue of $\tau_{\text{conn}} = 500 \gg \tau_{\text{cont}} = 10$.

\subsection{Symbolic Prior Lexicons}

\paragraph{Strong signal set} $\mathcal{R}_{\text{strong}}$: definitive author-labeled irony hashtags and their close orthographic variants:
\begin{quote}
\texttt{\#sarcasm, \#sarcastic, \#irony, \#ironic, \#not, \#justkidding, \#jk}
\end{quote}
Presence of any element assigns $S_{\text{prior}}(x) = \rho_{\text{strong}} = 0.88$.

\paragraph{Weak signal set} $\mathcal{R}_{\text{weak}}$: soft hashtag signals often co-occurring with ironic intent but insufficient alone:
\begin{quote}
\texttt{\#lol, \#sure, \#totally, \#obviously, \#right, \#yeah, \#great, \#awesome, \#fantastic, \#wonderful, \#perfect, \#love, \#amazing}
\end{quote}
Each contributes $\rho_{\text{weak}} = 0.15$ to the composite score.

\paragraph{Emoji contrast rules} $\mathcal{R}_{\text{emoji}}$: structural patterns where a positive-valence emoji is immediately followed by a negative-valence emoji or vice versa within the same token window. Positive cluster includes smiling, laughing, heart, thumbs-up, and celebration symbols. Negative cluster includes crying, angry, broken-heart, and eye-roll symbols. A match contributes $\rho_{\text{emoji}} = 0.25$.

\paragraph{Elongation tokens} $\mathcal{R}_{\text{elong}}$: tokens where any single character is consecutively repeated $\ge \eta = 3$ times (e.g., \textit{sooooo, yesssss, reallyyyy}). Detected via the regex \verb|(.)\1{2,}|. Contributes $\rho_{\text{elong}} = 0.12$.

\subsection{Negation-Aware Semantic Scan Vocabulary}
\label{app:vocabulary}

\paragraph{Ironic indicator tokens}: \\ \texttt{sarcasm, sarcastic, ironic, irony, clearly, obviously, totally, definitely, absolutely, sure, right, great, wonderful, perfect, love, exactly, certainly, naturally}.

\paragraph{Literal indicator tokens}: \\ \texttt{literal, sincere, genuine, actually, truly, honestly, factual, real, straightforward, serious, earnest, authentic}.

\paragraph{Local negation tokens} (invert the score of the subsequent indicator): \\ \texttt{not, no, never, without, hardly, barely, cannot, can't, won't, don't}.

\section{Chain-of-Thought Generation and Compression}
\label{app:cot}
\subsection{Few-Shot Chain-of-Thought Prompt Template}
\label{app:cot_prompt}
The prompt fed to the LLM for the chain-of-thought generation is as follows: 

\begin{quote}
\ttfamily
You are an expert linguist specializing in detecting irony and sarcasm in social media text.

\textbf{DEFINITION}: A tweet is IRONIC if there is a contrast between its literal meaning and its intended meaning, or if the author says the opposite of what they actually mean (often to mock, criticize, or be humorous). A tweet is NON-IRONIC if it is a sincere, literal statement.

\textbf{KEY SIGNALS TO CHECK:}
  - Does the literal meaning contradict the real-world situation?
  - Is there an exaggerated, over-the-top positive/negative tone?
  - Are there hashtags like \#not, \#sarcasm, \#irony, \#obviously that signal ironic intent?
  - Does the tweet mock or criticize something by pretending to praise it?
  - Would a reasonable reader take this at face value, or detect a hidden meaning?
  - Are there elongated words like "Loooove" or "Soooo" used sarcastically?

\textbf{CONFIDENCE SCALE:}
  0.0 = Absolutely certain NON-IRONIC (sincere, literal, no ambiguity at all);
  0.1 = Very likely non-ironic, tiny residual doubt;
  0.3 = Probably non-ironic, some mixed signals present;
  0.5 = Completely uncertain — could genuinely be either;
  0.7 = Probably ironic, some mixed signals present;
  0.9 = Very likely ironic, tiny residual doubt;
  1.0 = Absolutely certain IRONIC (clear sarcasm/irony, no ambiguity at all).

\textbf{EXAMPLES:}

\textbf{Tweet:} ``Oh great, another Monday. Just what I needed." \\
\textbf{Reasoning:} ``Just what I needed" is exaggeratedly positive about something universally disliked. Classic sarcasm with no ambiguity.
Score: 0.95.

\textbf{Tweet:} ``Happy birthday to my best friend! Hope your day is amazing." \\
\textbf{Reasoning:} Sincere, literal birthday wish. No hidden meaning, no contrast, tone matches content perfectly.
Score: 0.05.

\textbf{Tweet:} ``Wow, love how my flight got cancelled on the day of my interview. Truly blessed." \\
\textbf{Reasoning:} ``Truly blessed" after describing a disaster is a clear ironic inversion. Very high confidence.
Score: 0.92.

\textbf{Tweet:} ``This weather is something else today." \\
\textbf{Reasoning:} Ambiguous — could be genuine admiration or sarcastic complaint depending on context not available in the tweet alone.
Score: 0.50.

Now analyze the following tweet using the same reasoning process.

\textbf{Tweet:} \{\texttt{tweet}\}.

Think step by step through the KEY SIGNALS above. End your response with EXACTLY:
\textbf{Score:} X.XX
(a number between 0.00 and 1.00, two decimal places).
\end{quote}

\subsection{Role of Compression and Difference with TokenSkip}
\paragraph{Role of compression.}
Our framework employs a two-stage compression process. First, the LLM generation is bounded by a fixed budget of \texttt{max\_new\_tokens=120}, which can truncate reasoning before the LLM reaches its complete inference or explicit confidence score. Second, a post-generation pruning mechanism selectively retains the most downstream-relevant tokens from the generated sequence. Thus, the compressed CoT is constructed from whatever reasoning is available within the generation budget, rather than necessarily from the LLM's complete reasoning trajectory. When the LLM's explicit numerical score is pruned or if it is not available at all within the original reasoning length, we compute a new CoT-based confidence estimate from the surviving reasoning. Consequently, the resulting $P_{\mathrm{CoT}}$ is a compression-derived estimator and need not coincide with the LLM's original confidence. This distinction, and the motivation for discarding the original score, are discussed in in the following subsection.

\paragraph{Difference from TokenSkip.}
Here, the second stage of compression is performed through \textbf{post-generation token-pruning}, whereas in frameworks like TokenSkip, compression is learned during supervised fine-tuning and performed directly during autoregressive generation.

\subsection{Motivation for Discarding Pruned Confidence Scores}
\label{app:motivation}
It was observed that the restricted CoT rarely ended with a numerical score, Therefore in most cases, scores had to be computed via the methods described in subsection~\ref{score}. However, even if the score was computed, that was often discarded, as it was rarely classified as critical by the gradient-based scoring module. 

When the token encoding the LLM's explicit confidence score is pruned, its numerical prediction is discarded rather than retained as an independent signal. This design follows directly from the gradient-based definition of token relevance. For a token embedding $e_{c_i}$, a first-order perturbation
gives
\begin{equation}
\Delta \mathcal{L}_{\mathrm{CE}}
\approx
(\nabla_{e_{c_i}}\mathcal{L}_{\mathrm{CE}})^{\top}\Delta e_{c_i}.
\end{equation}
Thus, a token with low gradient sensitivity has limited local influence on the downstream classification objective. Retaining its explicit numerical score after pruning would allow information associated with a token deemed task-irrelevant to bypass the compression mechanism. We therefore reconstruct $P_{\mathrm{CoT}}$ from the surviving reasoning, maintaining consistency between token selection and the resulting CoT-derived signal.

This design choice does not imply that the raw LLM score is intrinsically uninformative. Rather, it ensures that the compressed CoT signal is derived
exclusively from the evidence retained by the task-aware selector. 

\subsection{Two-Stage CoT Compression and Confidence Reconstruction}
\label{app:compression_confidence}

RDS Fusion applies CoT compression in two stages:

\begin{enumerate}
    \item \textbf{Generation-time token constraint.} \texttt{Qwen2.5-3B-Instruct} generates the reasoning trajectory with \texttt{max\_new\_tokens}=120. This is a hard upper bound, not a target length, and therefore limits autoregressive generation cost. A trajectory may consequently terminate before the model completes its reasoning or produces an explicit confidence score.

    \item \textbf{Post-generation pruning.} The generated trajectory is subsequently processed by the Local Guardian. Tokens with insufficient downstream sensitivity are removed while preserving the order of retained tokens, producing the compressed trajectory $\tilde{c}$ used by the fusion layer.
\end{enumerate}

These two stages have different computational roles: the generation-time constraint bounds autoregressive computation, whereas post-generation pruning reduces the representation passed to downstream components. The latter cannot retroactively reduce the cost of tokens that have already been generated.

We distinguish the LLM's original confidence from the confidence used by RDS. Let $P_{\mathrm{LLM}}$ denote an explicit numerical confidence expressed in an uncompressed LLM trajectory, when available, and let $P_{\mathrm{CoT}}$ denote the estimator extracted from the compressed trajectory $\tilde{c}$. In general,
\[
P_{\mathrm{CoT}}(\hat{y}=1|\tilde{c}_{\text{RDS}}) \neq P_{\mathrm{LLM}}(\hat{y}=1|c).
\]

Formally, given the compressed trajectory $\tilde{c}_{\mathrm{RDS}}$, the following steps are followed sequentially to obtain $P_{\mathrm{CoT}}(\hat{y}=1|\tilde{c}_{\mathrm{RDS}})$:

\begin{enumerate}
    \item \textbf{Explicit score:} search $\tilde{c}_{\mathrm{RDS}}$ for
    $\texttt{Score: }s$, and set
    \[
    P_{\mathrm{CoT}}(\hat{y}=1|\tilde{c}_{\mathrm{RDS}})=\operatorname{clip}(s,0,1).
    \]

    \item \textbf{Verdict fallback:} if no score exists, parse a $\texttt{Verdict:}$ statement. Set $P_{\mathrm{CoT}}(\hat{y}=1|\tilde{c}_{\mathrm{RDS}})=0.9$ for an
    ironic/sarcastic verdict and $P_{\mathrm{CoT}}(\hat{y}=1|\tilde{c}_{\mathrm{RDS}})=0.1$ for a
    non-ironic/literal verdict.

    \item \textbf{Semantic fallback:} if neither is found, scan the final three sentences of $\tilde{c}_{\mathrm{RDS}}$. Let $r_i$ and $r_l$
    denote the weighted counts of ironic and literal indicators,
    respectively. Then
    \[
    P_{\mathrm{CoT}}(\hat{y}=1|\tilde{c}_{\mathrm{RDS}})
    =0.15+\frac{0.7r_i}{r_i+r_l},
    \] 
    clipped to $[0.15,0.85]$.

    \item \textbf{Full-text fallback:} if the semantic scan yields no indicators, apply the same mapping to the full $\tilde{c}_{\mathrm{RDS}}$.

    \item \textbf{Forced verdict:} if $P_{\mathrm{CoT}}(\hat{y}=1|\tilde{c}_{\mathrm{RDS}})=0.5$, reprompt the LLM using the compressed trajectory and a 10-token forced-verdict template, then repeat Steps 1--4.
\end{enumerate}

\paragraph{Computation of Semantic Indicator Scores.}
Let $\mathcal{I}$ and $\mathcal{L}$ denote the ironic and literal indicator vocabularies defined in Appendix~\ref{app:vocabulary}. For each occurrence $w$ of an indicator in the selected CoT span, we assign a context-sensitive weight
\[
V(w)=
\begin{cases}
3, & \text{for a contextual conclusion cue},\\
2, & \text{otherwise}.
\end{cases}
\]
Thus, each indicator occurrence receives a context-sensitive weight: $V(w)=3$ in a concluding context and $V(w)=2$ otherwise. The weighting is applied independently to each indicator; hence, ``\emph{clearly ironic}'' contributes $3+3=6$ when both indicators occur in a concluding context. Local negation reverses the polarity, so a negated ironic indicator contributes to $r_l$, while a negated literal indicator contributes to $r_i$. The resulting weighted scores are therefore
\[
r_i=\sum_{w\in\mathcal{I}}V(w),
\qquad
r_l=\sum_{w\in\mathcal{L}}V(w).
\]
If $r_i=r_l=0$, no semantic indicator is present and the pipeline enters into full-text fallback mechanism.

\subsection{Unrestricted Reasoning Ablation}
\label{app:unrestricted_reasoning}

To evaluate whether allowing the LLM to generate substantially longer reasoning improves classification performance, we conducted an unrestricted-CoT ablation using the same \texttt{Qwen2.5-3B-Instruct} model and downstream RoBERTa classifier, but without post-generation CoT compression. The generation budget was increased to \texttt{max\_new\_tokens}=1000 to allow the reasoning trajectory to evolve to its full extent. This value is nevertheless still a hard upper bound, and trajectories reaching 1000 tokens remain truncated. In practice, the generated unrestricted CoTs averaged approximately 200--500 tokens, substantially longer than the 120-token limit imposed in RDS Fusion before generation. The larger generation budget consequently incurred a considerably higher autoregressive computational cost.

\subsubsection{Accuracy and Latency}
On the complete 734-example evaluation set, unrestricted CoT with \texttt{max\_new\_tokens}=1000 achieved 74.9\% accuracy, compared with 78.1\% for RDS Fusion under the 120-token constraint. Thus, substantially increasing the reasoning budget did not improve overall performance. RDS Fusion required 2.8 hour to complete evaluation on the whole testset, whereas the ablation took 7.6 hour. 

\subsubsection{No Recallibration of $\alpha$}
It was also observed that for length-unrestricted reasonings, the entropies were larger than the case when the CoT was restricted to a threshold, i.e. the compressed case. This is expected from the definition of Shannon entropy. However, we did not re-calibrate $\alpha$ for this experiment. It was also observed that for this case, the range of variation entropy was actually quite large. However since the LLM almost always gave one sided / binary predictions (e.g., $0.00, 0.05, 0.92, 0.95$ etc), with estimations of more than $0.8$ in most of the cases, re-calibrating to increase $\alpha$ would only have made the performance worse. This is supported by the fact that there were only \textbf{8 false negatives} for this ablation.

\subsubsection{Precision--Recall Analysis}
The precision--recall behavior further illustrates the distinction between the three configurations. Table~\ref{tab:precision_recall} compares the ironic-class precision and recall of the RoBERTa-only baseline, the unrestricted CoT ablation, and RDS Fusion.

\begin{table}[h]
\centering
\begin{tabular}{lcc}
\toprule
\textbf{Method} & \textbf{Precision} & \textbf{Recall} \\
\midrule
RoBERTa-only            & 0.700 & 0.557 \\
Unrestricted CoT        & 0.614 & 0.979 \\
RDS Fusion              & 0.684 & 0.824 \\
\bottomrule
\end{tabular}
\caption{Ironic-class precision and recall across the baseline, unrestricted CoT ablation, and RDS Fusion.}
\label{tab:precision_recall}
\end{table}

RoBERTa-only exhibits relatively high precision but substantially lower recall, whereas unrestricted CoT achieves very high recall at the cost of markedly reduced precision. RDS Fusion occupies an intermediate regime, recovering most of the precision of RoBERTa while retaining substantially higher recall than the baseline. Thus, the compressed reasoning signal, when incorporated through the RDS Fusion mechanism, mitigates the strong recall bias observed in unrestricted CoT while avoiding the low-recall behavior of the RoBERTa-only configuration.

\subsection{Analysis of the Compressed CoT}
Upon analysis of the $5$ critical tokens across multiple cases, it was revealed that extracted critical tokens predominantly consist of the LLM’s generated reasoning vocabulary (e.g., ``contradiction", ``mocking"), however tokens from the raw tweets are also present in some cases. This confirms the module functions as a task-aware Chain-of-Thought distiller, isolating the LLM's core pragmatic analysis to feed the fusion layer. Furthermore, it was also found that the original score given by the LLM in its uncompressed CoT, was frequently discarded and through the mechanisms described in subsection \ref{score}, a new score was generated from the compressed CoT. 

\section{Additional Results}
\subsection{CoT and Adaptive Fusion Diagnostics}
\label{app:cot_fusion_diagnostics}

\begin{figure}[h]
\centering
\includegraphics[width=\linewidth]{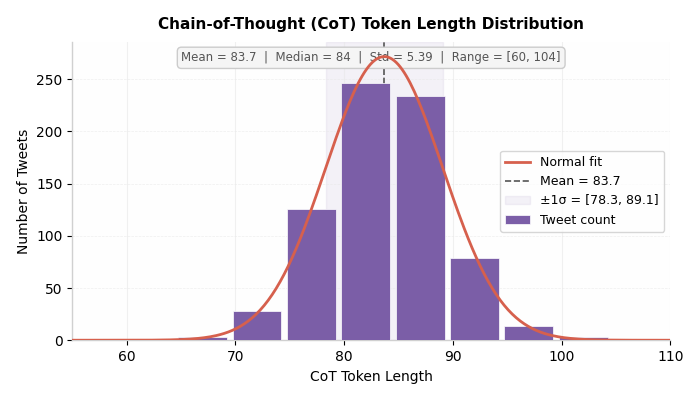}
\caption{Distribution of budget-limited CoT token lengths before post-generation pruning. Mean = 83.7, $\sigma=5.39$, with lengths ranging from 60--104 tokens.}
\label{fig:cot_dist}
\end{figure}

Figure~\ref{fig:cot_dist} shows the distribution of budget-limited CoT lengths \textbf{before post-generation pruning}. The lengths are tightly concentrated, with a mean of 83.7 tokens and standard deviation of 5.39, and remain below the 120-token generation limit.

\begin{figure}[h]
\centering
\includegraphics[width=\linewidth]{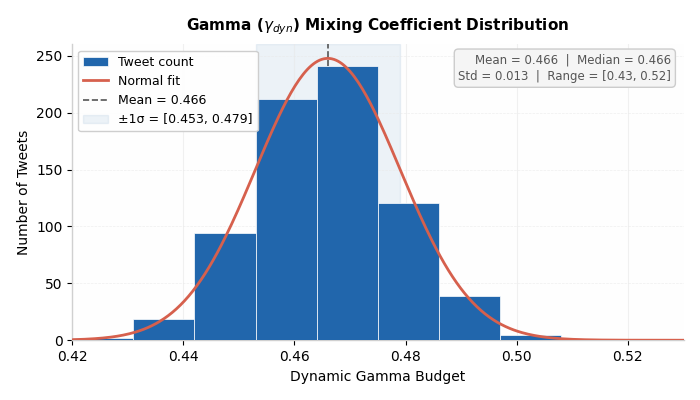}
\caption{Distribution of $\gamma_{\text{dyn}}$, which exhibits limited variation across instances ($\sigma=0.013$).}
\label{fig:gamma_dist}
\end{figure}

Figure~\ref{fig:gamma_dist} shows the distribution of $\gamma_{\text{dyn}}$. The small standard deviation ($\sigma=0.013$) indicates limited variation across instances.

\begin{figure}[h]
\centering
\includegraphics[width=\linewidth]{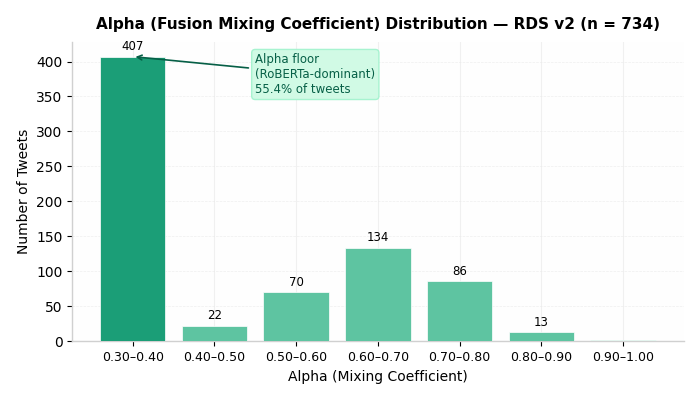}
\caption{Distribution of the adaptive fusion weight $\alpha$. The $[0.30,0.40]$ range contains 55.4\% of inferences, while $\alpha>0.5$ occurs in 41.3\%.}
\label{fig:alpha_dist}
\end{figure}

Figure~\ref{fig:alpha_dist} shows the distribution of the adaptive fusion weight $\alpha$. The lower bound $[0.30,0.40]$ is reached in 55.4\% of inferences, while $\alpha>0.5$ occurs in 41.3\% of cases.

\subsection{Symbolic Prior Activation}
\label{app:prior_activation}

\begin{figure}[h]
\centering
\includegraphics[width=\linewidth]{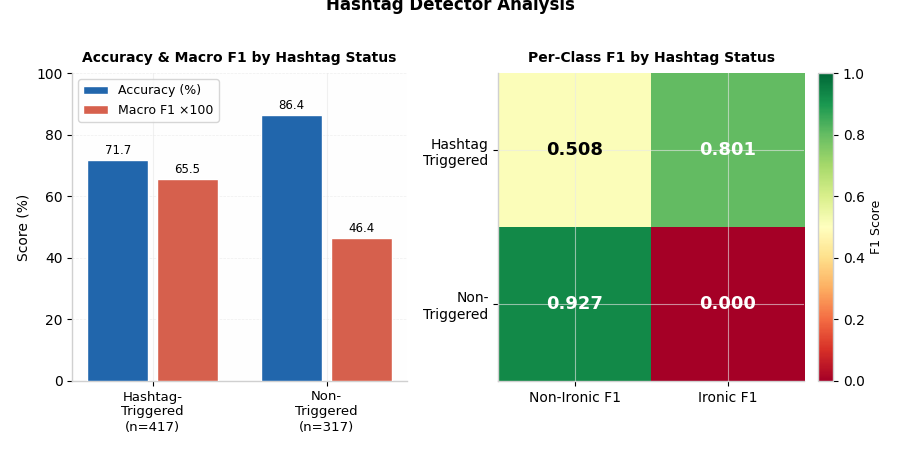}
\caption{Hashtag activation analysis across TweetEval and iSarcasm, illustrating the domain dependence of explicit syntactic cues.}
\label{fig:hashtag_analysis}
\end{figure}

Figure~\ref{fig:hashtag_analysis} compares the symbolic prior activation across TweetEval and iSarcasm. The difference in activation rates illustrates the domain dependence of these explicit syntactic cues.

\subsection{Cross-Domain Analysis}
\label{app:cross_domain}

\begin{figure}[h]
\centering
\includegraphics[width=\linewidth]{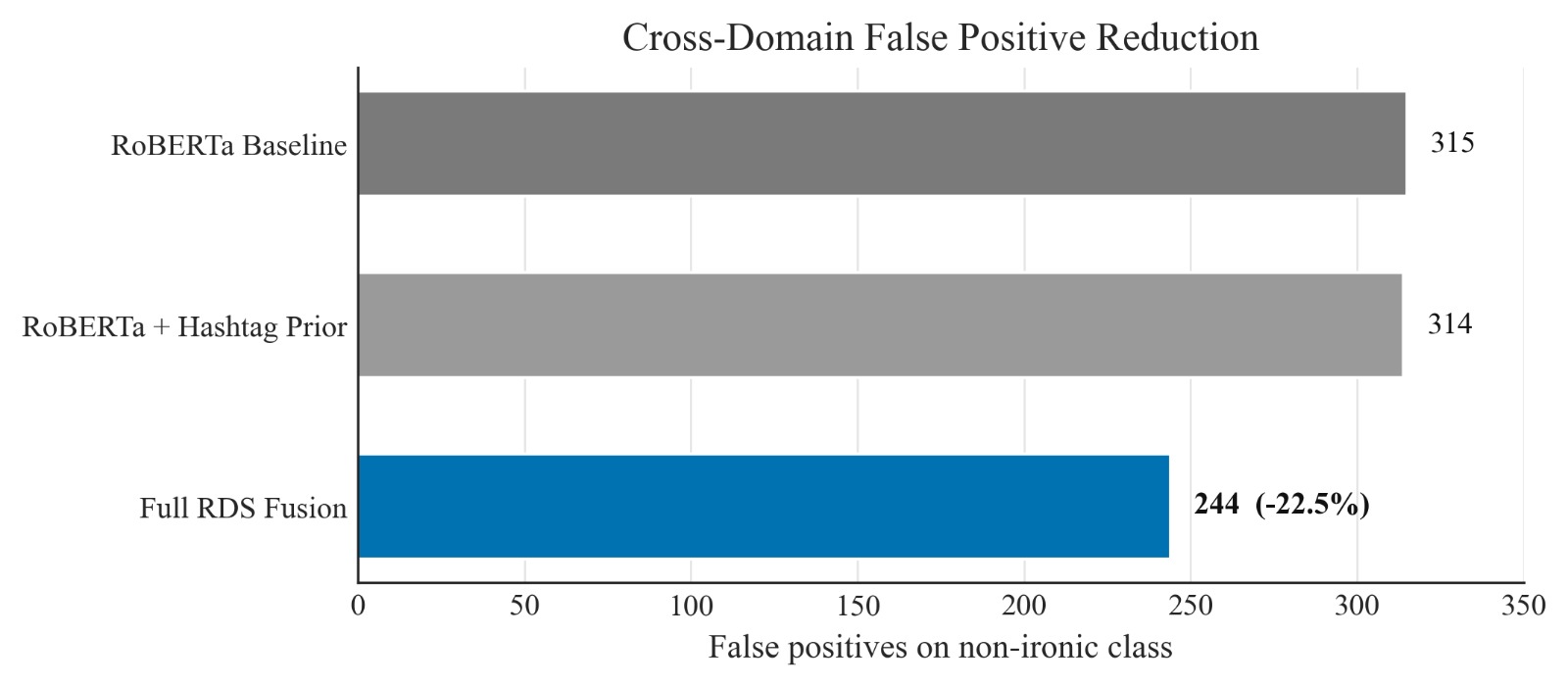}
\caption{Cross-domain false-positive reduction on iSarcasm. RDS Fusion reduces false positives from 315 to 244.}
\label{fig:isarcasm_fp}
\end{figure}

Figure~\ref{fig:isarcasm_fp} shows the reduction in false positives on iSarcasm, from 315 to 244.

\begin{figure}[h]
\centering
\includegraphics[width=\linewidth]{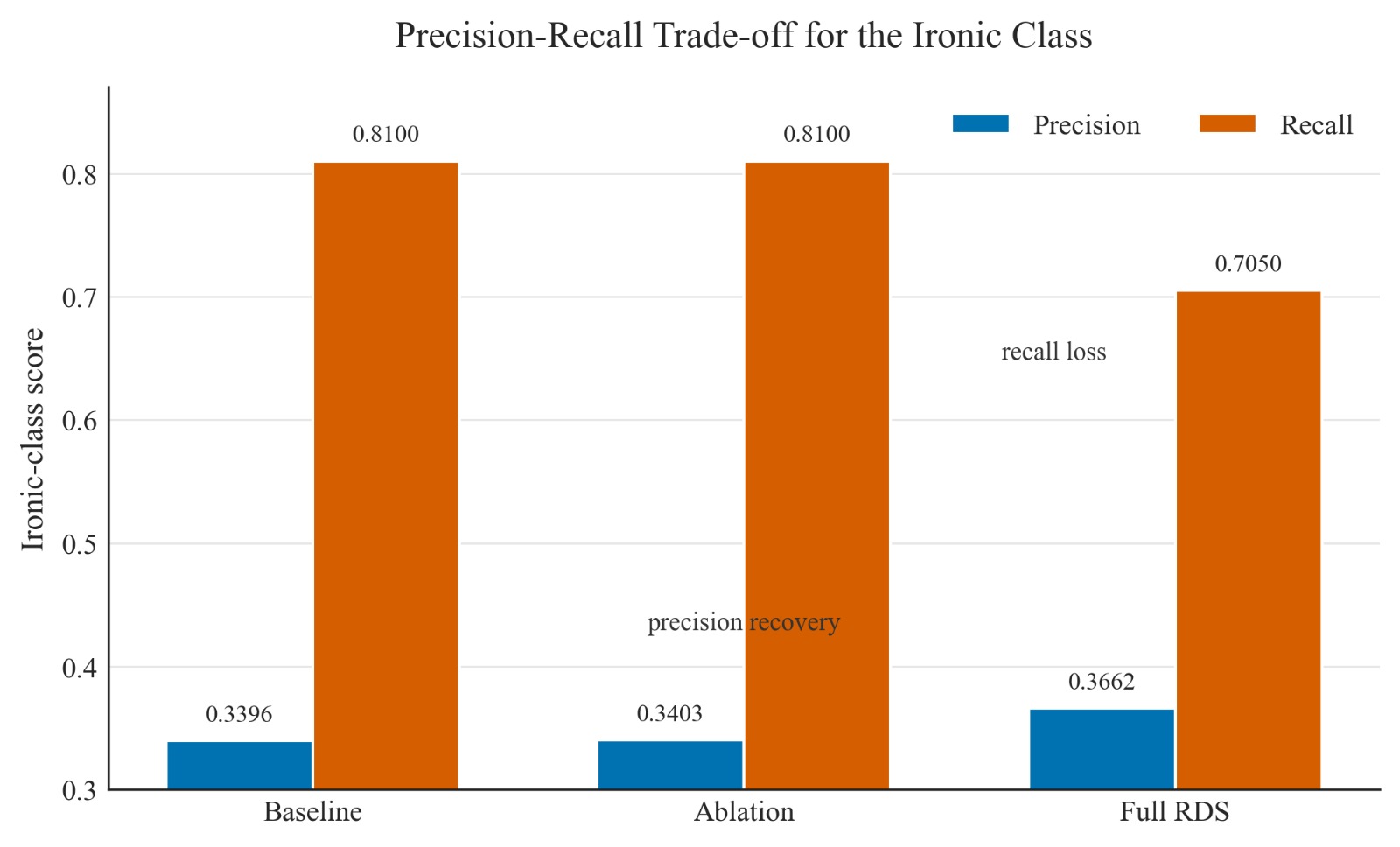}
\caption{Precision--recall trade-off on the iSarcasm ironic class, highlighting precision recovery under cross-domain evaluation.}
\label{fig:isarcasm_pr}
\end{figure}

Figure~\ref{fig:isarcasm_pr} shows the corresponding precision--recall trade-off for the ironic class, highlighting the precision-recovery behavior of RDS.

\section{Two More Case Studies}
\label{app:case_study}
\subsection{Case Study 2: Correct Irony Classification with Composite Signals (Tweet 172, TweetEval Dataset)}

\paragraph{Tweet:} \textit{Yay for Fire Alarms at 3AM \#not}

This tweet contains a strong explicit irony marker (\texttt{\#not}), which the symbolic prior immediately identifies, assigning $S_{\text{prior}}(x) = \rho_{\text{strong}} = 0.88$. This caps the baseline trust weight at $\alpha \leftarrow \min(\alpha, 0.30)$.

\paragraph{RoBERTa baseline:} The fine-tuned encoder independently assigns a very high irony probability ($P_{\text{RoBERTa}}(\hat{y}=1|x) = 0.980$), consistent with the overtly positive framing (``Yay'') applied to an objectively unpleasant event (a 3AM fire alarm). Both signals are strongly aligned.

\paragraph{CoT pipeline:} The LLM reasoning trajectory outputs $P_{\text{CoT}}(\hat{y}=1|\tilde{c}_{\text{RDS}}) = 0.85$, independently parsing the pragmatic incongruence between the enthusiastic exclamation and the disruptive context. With entropy $H(c) = 0.342$ mapping to $\gamma_{\text{dyn}} = 0.469$, the Local Guardian whitelists critical tokens including \textit{indicates}, \textit{disruptive}, \textit{excitement}, and \textit{Tone}. The strong signal override recomputes $P_{\text{base}}(\hat{y}=1|x) = 0.30 \times 0.980 + 0.70 \times 0.85 = 0.889$, and the final blend yields $P(\hat{y}=1|x) = 0.60 \times 0.889 + 0.40 \times 0.88 = 0.885$. Correct Ironic prediction.

This instance demonstrates the framework's most coherent operating mode: the symbolic prior, the RoBERTa encoder, and the CoT pipeline are unanimously aligned, producing a high-confidence, correctly calibrated ironic prediction.

\subsection{Case Study 3: Failure under Conflicting Signals (Tweet 143, TweetEval Dataset)}

\paragraph{Tweet:} \textit{how is it possible that somebody so interesting is also so, so, so boring? \#irony \#what}

This is a ground-truth ironic tweet containing a definitive author-labeled marker (\texttt{\#irony}). The symbolic prior correctly activates on this strong signal, assigning $S_{\text{prior}}(x) = 0.88$ and engaging the strong-signal override to cap the baseline trust weight at $\alpha = 0.300$.

\paragraph{RoBERTa baseline:} The fine-tuned encoder assigns a severely low irony probability ($P_{\text{RoBERTa}}(\hat{y}=1|x) = 0.087$), entirely failing to resolve the self-contradictory predication (``interesting'' and ``boring'' applied to the same subject).

\paragraph{CoT pipeline:} The compressed CoT output is $P_{\text{CoT}}(\hat{y}=1|\tilde{c}_{\text{RDS}}) = 0.15$ with a predictive entropy of $H(c) = 0.249$. The reasoning trajectory misinterprets the oxymoronic construction as a literal complaint, lacking the contextual agility to recognize the pragmatic inversion.

\paragraph{Adaptive Fusion:} Because both neural components fail with extreme confidence, their combined base probability collapses to $P_{\text{base}}(\hat{y}=1|x) = 0.30 \times 0.087 + 0.70 \times 0.15 = 0.131$. Even after applying the heavy strong-signal blend ($\omega_{\text{strong}} = 0.60$), the prior cannot rescue the prediction: $P(\hat{y}=1|x) = 0.60 \times 0.131 + 0.40 \times 0.88 = 0.431$. The result falls below the 0.5 decision threshold, yielding a false negative.

This case exposes a critical architectural boundary: the limits of the symbolic prior's rescue capacity. When both the fine-tuned encoder and the compressed CoT confidently fail in unison, their combined bias mathematically overpowers even a perfectly fired, high-confidence explicit marker.


\section{Comparison of Unrestricted, Restricted and Post-pruned CoT Trajectories}
\label{app:three_way_cot}

To provide a qualitative view of the two-stage reasoning constraint, we examine a single held-out example (Tweet \#51) under three inference configurations: unrestricted CoT, CoT restricted by the 120-token budget, and post-generation Local Guardian pruning of the restricted trajectory. All three configurations use the same manually authored prompt. The unrestricted configuration permits up to 1000 generated tokens and directly extracts the final score produced by the LLM, whereas the restricted and post-pruned configurations use the confidence-reconstruction procedure described in subsection~\ref{score}. 

\paragraph{Tweet:} \textit{Rangers league game with Alloa moved because of the Petrofac Cup final in April... \#irony.}\\
\textbf{True Label:} $y=1$ (ironic).

\begin{table}[h]
\centering
\begin{tabular}{lcc}
\toprule
\textbf{Configuration} & \textbf{Length (words)} & \textbf{Score} \\
\midrule
Unrestricted CoT & 321 & 0.05 \\
Restricted CoT & 86 & 0.69 \\
Post-pruned CoT & 47 & 0.57 \\
\bottomrule
\end{tabular}
\caption{Three-way comparison of CoT trajectories for Tweet \#51. The unrestricted configuration reports the raw LLM score, while the restricted and post-pruned configurations report reconstructed CoT signals.}
\label{tab:three_way_cot}
\end{table}

\subsection{Unrestricted CoT}
To illustrate the effect of the two-stage compression process, we first examine the model's unrestricted reasoning trajectory. With unrestricted generation, the model produced the following 321-word reasoning trajectory and assigned a raw LLM score of \textbf{0.05}:

\vspace{0.5 em}

\noindent
\hspace*{0.05\columnwidth}%
\parbox{0.90\columnwidth}{%
\textbf{CoT Generated by the LLM with \texttt{max\_new\_tokens = 1000}}%
}

\begin{center}
\small
\textbf{Highlighting:}
\colorbox{yellow!25}{Semantic evidence} \,
\colorbox{black!10}{Decisive reasoning}
\end{center}

\begin{mdframed}[
  leftmargin=0.05\columnwidth,
  rightmargin=0.05\columnwidth
]
\small
Let's analyze the given tweet step by step:

\textbf{Tweet Analysis:}

\textbf{Tweet:} `Rangers league game with Alloa moved because of the Petrofac Cup final in April... \#irony'

\textbf{Step-by-Step Reasoning:}

1. \textbf{Literal Meaning vs. Real-World Situation:}
   - The literal meaning of the tweet is that a Rangers league game against Alloa was rescheduled due to the Petrofac Cup final.
   - \hl{There is no apparent contradiction between the literal meaning and the real-world situation.} The tweet is stating a factual event without any indication of irony.

2. \textbf{Exaggerated Tone:}
   - The tweet does not use exaggerated language or phrases that would suggest irony.
   - \hl{Phrases like ``moved because of the Petrofac Cup final in April'' are straightforward and do not appear overly enthusiastic or negative.}

3. \textbf{Hashtags:}
   - The hashtag ``\hl{\#irony}'' is present, which might suggest that the author intends for the tweet to be interpreted as ironic.
   - However, \hl{the presence of a hashtag alone is not enough to confirm irony;} it often indicates a deliberate attempt to convey irony but doesn't always work.

4. \textbf{Mockery/Criticism:}
   - There is no clear mockery or criticism in the tweet.
   - \hl{The author is simply reporting a fact about a rescheduled game.}

5. \textbf{Reader Interpretation:}
   - While a reasonable reader might interpret this as ironic, \hl{the lack of additional context or explicit indicators makes it difficult to confidently assert that the tweet is ironic.}

6. \textbf{Word Usage:}
   - The tweet uses common phrases (``moved because of'') without any elongated or exaggerated words that typically accompany irony.

\textbf{Conclusion:}

\colorbox{black!10}{%
\begin{minipage}{0.93\linewidth}
The tweet is straightforward and does not contain any clear indicators of irony. It is simply reporting a factual event without any apparent contradiction or exaggerated tone.
\end{minipage}
}

\textbf{Score:}
\textbf{Score: 0.05}

This score reflects the very low likelihood that the tweet is ironic, given the absence of clear indicators and the straightforward nature of the information provided. The tweet is most likely a sincere statement about a rescheduled game.
\end{mdframed}

\vspace{0.5 em}

The unrestricted trajectory explicitly recognizes the \#irony marker but continues with substantially longer deliberation about whether the marker should be trusted, ultimately producing a low irony score. This example illustrates that a larger reasoning budget can lead to a materially different confidence assignment.

\subsection{Restricted CoT}
Applying the 120-token generation ceiling used by RDS Fusion produces the following 86-word trajectory:

\vspace{0.5 em}

\noindent
\hspace*{0.05\columnwidth}%
\parbox{0.90\columnwidth}{%
\textbf{CoT Restricted by \texttt{max\_new\_tokens = 120}.}
\quad
Tokens marked \retained{Green} are \retained{Retained} after post-pruning;
tokens marked \deleted{Red} are \deleted{Deleted}.
}
\begin{mdframed}[
  leftmargin=0.05\columnwidth,
  rightmargin=0.05\columnwidth
]
\small
\deleted{To} \retained{analyze} \deleted{the} \retained{given tweet, let's break} \deleted{it} \deleted{down} \deleted{using} \retained{the} \deleted{provided} \retained{key signals:}

\textbf{\retained{Key Signals Analysis:}}

1. \textbf{\retained{Literal Meaning}} \deleted{vs.} \textbf{\retained{Intended Meaning:}}

\deleted{- The} \retained{tweet states:} ``\retained{Rangers league game} \deleted{with} \retained{Alloa moved} \deleted{because of the} \retained{Petrofac Cup final} \deleted{in} \retained{April... \#irony}''

\deleted{- The} \retained{literal meaning} \deleted{is that a} \retained{Rangers game} \deleted{against} \retained{Alloa} \deleted{was} \retained{rescheduled due} \deleted{to the} \retained{Petrofac Cup final.}

\deleted{- The} \retained{intended meaning,} \deleted{as} \retained{indicated} \deleted{by the} \retained{hashtag \#irony,} \retained{suggests} \deleted{that the} \retained{reason} \deleted{for the} \retained{rescheduling} \deleted{is} \retained{ironic} \deleted{or} \retained{not} \deleted{what it seems.}
\end{mdframed}

\vspace{0.5 em}

The restricted trajectory terminates before the extended reconsideration present in the unrestricted generation and yields a reconstructed CoT signal of \textbf{0.69}. The entropy of this sequence was found to be $0.24$ and therefore $\gamma_{\text{dyn}}$ was $0.448$. 

\subsection{Post-Pruned CoT}
Local Guardian is then applied to the restricted trajectory using the target-specific retention rule described in subsection~\ref{grad}. Therefore, a pruned chain-of-thought compressed by the ratio $\gamma_{\text{dyn}} (=0.448$ in this case) is obtained. The resulting 47-word trajectory is:

\vspace{0.5 em}

\noindent
\hspace*{0.05\columnwidth}%
\parbox{0.90\columnwidth}{%
\textbf{Pruned CoT, Used by RDS Fusion: \retained{Retained} and \critical{Critical} Tokens}%
}
\begin{mdframed}[
  leftmargin=0.05\columnwidth,
  rightmargin=0.05\columnwidth
]
\small
\retained{analyze given tweet, let's break using provided key signals:} \textbf{\retained{Key Signals}} \critical{Analysis:} \textbf{\retained{Literal Meaning Intended Meaning:}} \retained{tweet states:} ``\retained{Rangers league game Alloa moved Petrofac Cup final April... \#irony}'' \retained{literal meaning Rangers game Alloa rescheduled due Petrofac Cup final.} \critical{intended}, \retained{meaning,} \critical{indicated} \retained{hashtag \#irony,} \critical{suggests} \critical{ironic} \retained{not}
\end{mdframed}

\vspace{0.5 em}

The post-pruned trajectory yields a reconstructed CoT signal of \textbf{0.57}. The critical tokens identified for this example are \critical{\texttt{ironic}}, \critical{\texttt{Analysis}}, \critical{\texttt{indicated}}, \critical{\texttt{suggests}}, and \critical{\texttt{intended}}. Notably, the retained trajectory continues to contain the literal-versus-intended-meaning distinction and the explicit ironic cue despite substantial token removal.

\subsection{Analysis}
Overall, the example illustrates how the successive transformations introduced by the inference pipeline can alter the semantic evidence available for classification. The ground-truth label of the tweet is ironic ($y=1$), whereas the unrestricted 321-word reasoning trajectory incorrectly interprets the tweet as non-ironic and assigns a raw confidence of only $0.05$. Applying the generation-time budget substantially shortens the trajectory to 86 words; although the explicit confidence information may be removed during this restriction, the surviving reasoning provides sufficient semantic evidence for confidence reconstruction, yielding a score of $0.69$, which is on the correct ironic side of the decision boundary. The subsequent Local Guardian performs a more targeted token-level reduction, retaining the most task-relevant tokens and reducing the trajectory further to 47 words. The reconstructed confidence remains above the decision threshold at $0.57$, thereby preserving the correct prediction despite the substantially shorter reasoning trajectory.

The example therefore illustrates two important effects of the proposed compression strategy. Compression does not merely reduce the length of the reasoning trajectory; it can also change the semantic evidence available for downstream confidence extraction. On this particular TweetEval test-set, unrestricted CoT yields the worst performance, marginally below the restricted trajectory and $2.6\%$ below the restricted and post-pruned CoT. The progression from $0.05$ to $0.69$ and subsequently to $0.57$ should not, however, be interpreted as evidence that shorter reasoning is always superior. Rather, it provides a qualitative example of how successive compression stages can remove distracting or misleading reasoning while retaining sufficient task-relevant evidence to recover and preserve the correct prediction.

\section{Hardware, Environment, and Reproducibility Details}
\label{app:hardware}

\subsection{Hardware Configuration}

All experiments were executed on Kaggle and Google Colab environments with NVIDIA dual T4 GPU acceleration (2$\times$16 GB VRAM, CUDA 12.2). The TweetEval evaluation ($N=734$) required approximately 2.8 hours. The 120-token generation ceiling prevents arbitrarily long autoregressive reasoning; post-generation pruning produces the compact trajectory used downstream.

\subsection{Generation Parsing Guardrails}

The following secondary guardrails were applied to CoT outputs and predictive calculations before final confidence extraction:

\begin{enumerate}
    \item \textbf{Encoding and Formatting Standardization:} Model generations are decoded with \texttt{skip\_special\_tokens=True} and standardized via lowercase conversion and whitespace stripping prior to parsing.
    \item \textbf{Score Boundary Clamping:} Any extracted continuous confidence score is mathematically bounded using a min-max clamp to ensure it falls within the valid $[0.0, 1.0]$ probability range.
    \item \textbf{Unresolvable Ambiguity Fallback:} If the hierarchical regex and semantic parsers fail to identify a definitive score within the generated text, the extraction naturally defaults to $P_{\text{CoT}}(\hat{y}=1|\tilde{c}_{\text{RDS}}) = 0.5$. This explicitly triggers the two-pass fallback mechanism, which re-prompts the model for a forced verdict.
    \item \textbf{Entropy Underflow Prevention:} During the predictive entropy $H(c)$ computation, an epsilon value ($\epsilon
    = 10^{-9}$) is added to the token probabilities prior to the logarithmic transformation to prevent undefined \texttt{log(0)} errors. A secondary \texttt{NaN} catch enforces a $0.0$ entropy baseline if precision limits are exceeded.
\end{enumerate}

\subsection{Quantization Details}

\texttt{Qwen2.5-3B-Instruct} was loaded using 4-bit NF4 (Normal Float 4) quantization via \texttt{bitsandbytes} to optimize memory footprint:
\begin{itemize}
    \item \textbf{Quantization type:} \texttt{nf4} (NormalFloat 4-bit, optimized for normally distributed weights \cite{yang2024qwen25}).
    \item \textbf{Double quantization:} enabled (quantizes the quantization constants to further reduce memory).
    \item \textbf{Compute dtype:} \texttt{float16}.
    \item \textbf{Full model (RDS Fusion) memory footprint (post-quantization):} $\approx$ 2.1 GB VRAM.
    \item \textbf{Generation (RDS Fusion):} greedy decoding (\texttt{temperature=0}, \texttt{do\_sample=False}), \texttt{max\_new\_tokens=120}.
    \item \textbf{Generation (Unrestricted Reasoning):} greedy decoding (\texttt{temperature=0}, \texttt{do\_sample=False}), \texttt{max\_new\_tokens=1000}.
    \item \textbf{Two-pass fallback:} same model, \texttt{max\_new\_tokens=10}.
\end{itemize}

\subsection{Validation Protocol and Anti-Leakage Guarantee}

To prevent data leakage, the first 50 tweets of the TweetEval test set were designated as a validation split \emph{before} any parameter search. All calibration of $H_{\text{min}}$, $H_{\text{max}}$, $\alpha_{\text{min}}$, $\alpha_{\text{max}}$, $\rho_{\text{strong}}$, and blending weights was strictly confined to these 50 samples (achieving an over-fitted validation accuracy of 88.0\% on this calibration split). The remaining 734 tweets were treated as a completely unseen held-out test set. The iSarcasm evaluation used frozen TweetEval calibrations with zero additional tuning, guaranteeing a strictly few-shot cross-domain evaluation.

\subsection{Consolidated Calibration Thresholds}
\label{hyperparameters}
Table \ref{tab:hyperparameters} consolidates all static thresholds and fusion parameters derived from the 50-tweet validation split.

\begin{table}[h!]
\centering
\resizebox{\columnwidth}{!}{
\begin{tabular}{llc}
\hline
\textbf{Module} & \textbf{Parameter} & \textbf{Value} \\
\hline
Compression Budget & $H_{\text{min}}, H_{\text{max}}$ & 0.2, 0.5 \\
Compression Bounds & $\gamma_{\text{min}}, \gamma_{\text{max}}$ & 0.1, 0.9 \\
Retention Overrides & $\tau_{\text{crit}}, \tau_{\text{conn}}$ & 1000, 500 \\
Retention Overrides & $\tau_{\text{cont}}, \tau_{\text{low}}$ & 10, 1 \\
Prior Weights & $\rho_{\text{strong}}, \rho_{\text{cap}}$ & 0.88, 0.70 \\
Weak Prior Weights & $\rho_{\text{weak}}, \rho_{\text{emoji}}, \rho_{\text{elong}}$ & 0.15, 0.25, 0.12 \\
Fusion Bounds & $\alpha_{\text{min}}, \alpha_{\text{max}}$ & 0.45, 0.90 \\
Skepticism Gate & $H_{\text{sk}}, \lambda_{\text{sk}}, p_{\text{sk}}$ & 0.25, 0.5, 0.15 \\
Blend Weights & $\omega_{\text{strong}}, \omega_{\text{weak}}$ & 0.60, 0.75 \\
\hline
\end{tabular}
}
\caption{Consolidated pipeline hyperparameters.}
\label{tab:hyperparameters}
\end{table}

\subsection{Library Versions}

\begin{table}[H]
\centering
\begin{tabular}{ll}
\hline
\textbf{Library} & \textbf{Version} \\
\hline
\texttt{transformers} & 4.40.0 \\
\texttt{torch} & 2.2.0+cu121 \\
\texttt{bitsandbytes} & 0.43.1 \\
\texttt{llmlingua} & 0.2.2 \\
\texttt{numpy} & 1.26.4 \\
\texttt{scikit-learn} & 1.4.2 \\
\texttt{statsmodels} & 0.14.1 \\
\hline
\end{tabular}
\caption{Library versions for reproducibility.}
\label{tab:library_versions}
\end{table}

\end{document}